\documentclass[10pt,twocolumn,letterpaper]{article}

\usepackage{float}
\usepackage[accsupp]{axessibility}  

\usepackage[pagenumbers]{cvpr} 

\definecolor{cvprblue}{rgb}{0.21,0.49,0.74}
\usepackage[pagebackref,breaklinks,colorlinks,allcolors=cvprblue]{hyperref}

\def\paperID{972} 
\def\confName{CVPR}
\def\confYear{2025}

\title{SPMTrack: \underline{S}patio-Temporal \underline{P}arameter-Efficient Fine-Tuning with \\ \underline{M}ixture of Experts for Scalable Visual Tracking}

\author{Wenrui Cai$^1$, Qingjie Liu$^{1,2,3,*}$, Yunhong Wang$^{1,3}$\\
$^1$State Key Laboratory of Virtual Reality Technology and Systems, Beihang University, Beijing, China  \\
$^2$Zhongguancun Laboratory, Beijing, China  \\
$^3$Hangzhou Innovation Institute, Beihang University, Hangzhou, China \\
{\tt\small \{wenrui\_cai, qingjie.liu, yhwang\}@buaa.edu.cn}
}

\begin{document}
\maketitle
\begin{abstract}
Most state-of-the-art trackers adopt one-stream paradigm, using a single Vision Transformer for joint feature extraction and relation modeling of template and search region images. 
However, relation modeling between different image patches exhibits significant variations. 
For instance, background regions dominated by target-irrelevant information require reduced attention allocation, while foreground, particularly boundary areas, need to be be emphasized.
A single model may not effectively handle all kinds of relation modeling simultaneously. 
In this paper, we propose a novel tracker called \textbf{SPMTrack} based on mixture-of-experts tailored for visual tracking task (TMoE), combining the capability of multiple experts to handle diverse relation modeling more flexibly.
Benefiting from TMoE, we extend relation modeling from image pairs to spatio-temporal context, further improving tracking accuracy with minimal increase in model parameters.
Moreover, we employ TMoE as a parameter-efficient fine-tuning method, substantially reducing trainable parameters, which enables us to train SPMTrack of varying scales efficiently and preserve the generalization ability of pretrained models to achieve superior performance.
We conduct experiments on seven datasets, and experimental results demonstrate that our method significantly outperforms current state-of-the-art trackers. 
The source code is available at \href{https://github.com/WenRuiCai/SPMTrack}{https://github.com/WenRuiCai/SPMTrack}.
\end{abstract}  
\renewcommand{\thefootnote}{}
\footnote{$^*$Corresponding author.}
\section{Introduction}
\label{sec:intro}
Visual tracking aims to predict the location of a target throughout subsequent frames of a video, given the initial state in the first frame template.
Currently, most state-of-the-art trackers adopt a Transformer-based \cite{Vaswani2017AttentionIA} one-stream paradigm \cite{ye_2022_joint, Cui_2022_MixFormer, Wu_2023_CVPR_dropmae, chen2022backbone, Cai_2024_CVPR_HIPTrack, ODTrack, LoRAT}, utilizing a single Vision Transformer \cite{dosovitskiy2020image} backbone that accepts both the template and the search region image as input, and employs self-attention to perform image feature extraction and template–search region relation modeling simultaneously.

However, not all relation modeling between image patches benefits tracking performance.
Prior studies \cite{ye_2022_joint, Gao_2023_CVPR_GRM, Cai_2023_ICCV_ROMTrack,Yang_2023_ICCV_FBDMTrack} have demonstrated that the abundance of background patches in the search region deteriorates the discriminative power of the foreground features, and single vanilla attention has limited capability in suppressing undesirable foreground-background interactions.
Meanwhile, there are also some kinds of relation modeling that demonstrate positive effects on tracking performance \cite{gao2022aiatrack,fu2022sparsett}. For instance, enhancing attention to target boundary regions can improve tracking performance \cite{fu2022sparsett}, especially in challenging scenarios such as motion blur, partial occlusion, and deformation.

\begin{figure}[t]
\centering
\includegraphics[scale=0.58]{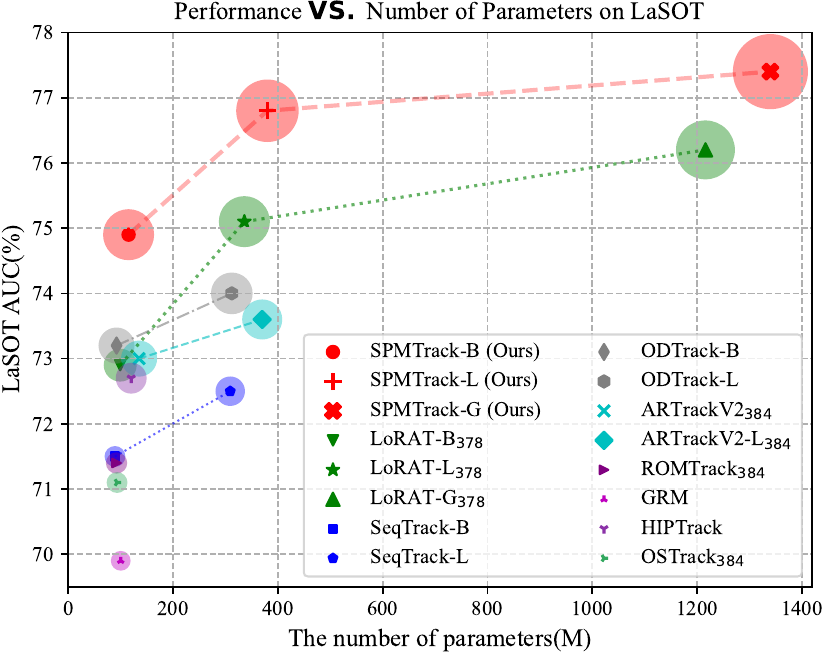}
\caption{Comparison of LaSOT AUC and model parameter count across different trackers. Larger loop indicates better performance.}
\label{fig:1}
\end{figure}

To address various relation modeling, 
numerous methods propose specific designs, including background filtering to reduce target-irrelevant information \cite{ye_2022_joint, Cai_2024_CVPR_HIPTrack}, customizing different interaction modules for tokens belonging to different categories \cite{Cai_2023_ICCV_ROMTrack, Gao_2023_CVPR_GRM}, and attention score adjustment based on foreground-background weights of each token \cite{Yang_2023_ICCV_FBDMTrack}. 
However, existing methods can only handle one or a few predefined types of relation modeling, and the hand-crafted strategies inherently lack adaptability.
Furthermore, there may exist many latent relationships that further complicate the specialized design. These limitations motivate our question: \textbf{\emph{How to design a tracker to adaptively process various relation modeling between image patches?}}






Inspired by mixture of experts (MoE) \cite{jacobs1991adaptive_moe,pmlr-v162-du22c_moe2,Aljundi_2017_CVPR_expert_gate,puigcerversparse_from_sparse_to_soft_moe,dai-etal-2024-deepseekmoe} in natural language processing, in this paper, we address the question by proposing \textbf{SPMTrack}, a novel tracker based on TMoE, a specialized mixture of experts module for visual tracking task. 
Similar to MoE, TMoE handles diverse relation modeling through adaptive weighted combinations of multiple experts. However, unlike traditional MoE that is exclusively applied to feed-forward network (FFN) layers, TMoE is simultaneously applied to the linear layers within both the attention layers and the FFN layers. This finer-grained application enables TMoE to achieve more diverse expert combinations, thereby enhancing the capability of various relation modeling. Moreover, differing from traditional MoE that uses FFNs as experts, 
all the experts in TMoE are linear layers and designed with a lightweight and efficient structure to ensure overall  efficiency.
Furthermore, we employ TMoE as a parameter-efficient fine-tuning method, which enables us to train SPMTrack of larger scales.
In SPMTrack, only a subset of experts and the router within TMoE, along with the prediction head need to be trained, which substantially reduces the trainable parameters while preserving the generalization capability of pretrained models. Notably, our method reduces the number of trainable parameters by nearly 80\% compared to \cite{Bai_2024_CVPR_ARTrackV2}, while demonstrates better performance.

Benefiting from the powerful capability of TMoE in handling diverse relation modeling, unlike previous one-stream trackers that only perform relation modeling between template-search region pairs, we extend SPMTrack to incorporate multi-frame spatio-temporal context modeling, which further enhances the performance of SPMTrack with minimal additional parameters.
We evaluated the performance of our method on seven datasets, and the experimental results demonstrate that SPMTrack achieves state-of-the-art performance across multiple datasets including LaSOT \cite{fan2019lasot}, GOT-10K \cite{huang2019got}, TrackingNet \cite{muller2018trackingnet} and TNL2K \cite{Wang_2021_CVPR_TNL2k}. Through extensive experiments with models of varying scales, as shown in Figure \ref{fig:1}, Table \ref{table:model_settings} and Table \ref{whole_comparison},  with ViT-B \cite{dosovitskiy2020image} as backbone and less than 30\% of parameters need to be trained, SPMTrack-B achieves or even surpasses the best performance of previous trackers utilizing ViT-L.

The main contributions of
this paper can be summarized as: \textbf{(1)} We propose TMoE, a mixture of experts module tailored for visual tracking. TMoE enhances the capability to handle various relation modeling and can be used as a method of parameter-efficient fine-tuning. \textbf{(2)} Based on TMoE, we propose SPMTrack, a novel tracker that can effectively integrate spatio-temporal context for visual tracking. \textbf{(3)} Experimental results demonstrate that SPMTrack achieves state-of-the-art performance across multiple datasets. We trained SPMTrack of varying scales, and our method with ViT-B as backbone can achieve or even surpass current trackers using ViT-L.


\section{Related Work}
\label{sec:related}

\subsection{One-Stream Trackers}


One-stream trackers employ a single Vision Transformer \cite{dosovitskiy2020image} as backbone, simultaneously performing feature extraction and relation modeling at each layer \cite{Xie_2022_Correlation,ye_2022_joint,chen2022backbone, Cui_2022_MixFormer}. Many efforts focus on incorporating historical context. TATrack \cite{he2023target_TATrack} and ODTrack \cite{ODTrack} propose architectures capable of processing multiple frames, while HIPTrack \cite{Cai_2024_CVPR_HIPTrack} introduces historical target features through prompt learning. SeqTrack \cite{Chen_2023_CVPR_seqtrack}, ARTrack\cite{Wei_2023_CVPR_autoregressive}, and ARTrackV2 \cite{Bai_2024_CVPR_ARTrackV2} achieve more precise predictions by incorporating multiple historical target position. Other works try to improve conventional self-attention to handle different relation modeling. OSTrack \cite{ye_2022_joint} progressively filters background regions to reduce foreground-background relation modeling, GRM \cite{Gao_2023_CVPR_GRM} and ROMTrack \cite{Cai_2023_ICCV_ROMTrack} categorizes image tokens and constrains relation modeling between specific categories, F-BDMTrack \cite{Yang_2023_ICCV_FBDMTrack} adjusts the attention score by calculating the foreground-background weights of tokens. But these methods can only handle specific relation modeling and rely on manually designed strategies.

\begin{figure*}[!t]
\centering
\includegraphics[scale=0.54]{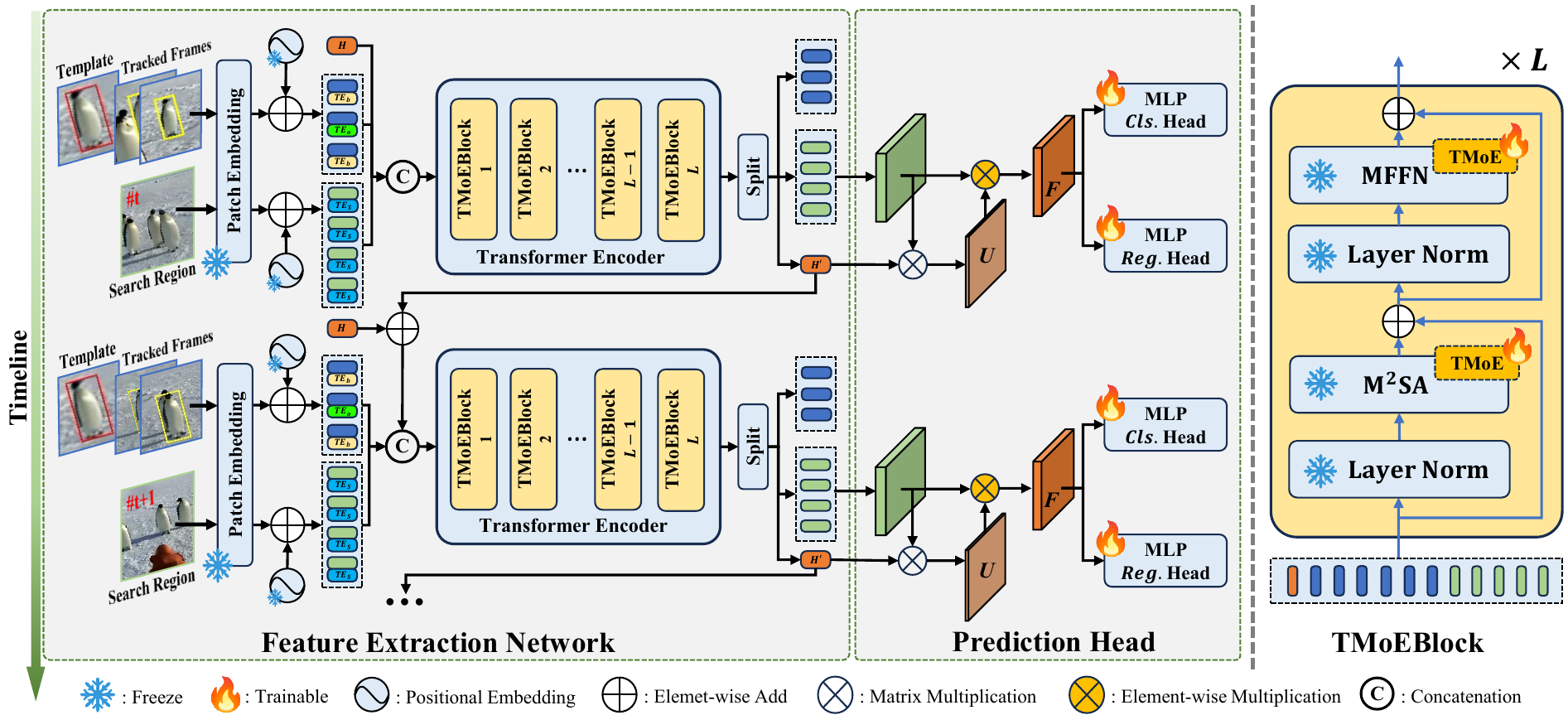}
\vspace{-1ex}
\caption{Overview of SPMTrack that consists of a feature extraction network and a prediction head. The main body of the feature extraction network is a Transformer encoder composed of multiple TMoEBlocks. The structure of TMoEBlock is shown on the right side. 
}
\label{fig:pipeline}
\end{figure*}

\subsection{Mixture of Experts}


Mixture-of-experts (MoE) \cite{jacobs1991adaptive_moe,pmlr-v162-du22c_moe2,Aljundi_2017_CVPR_expert_gate} are widely applied in large language models \cite{,puigcerversparse_from_sparse_to_soft_moe,dai-etal-2024-deepseekmoe,wumixture_mixture_of_lora_experts,ma2018modeling_multigate_moe,fedus2022switchtransformers,clark2022unified_scaling_laws_for_routed}, where each expert focuses on specific aspects of the data or particular tasks \cite{wang2024survey_data_synthesis_ours}. 
By performing an adaptive weighted sum over multiple experts, MoE can better capture complex patterns and relationships within the input data. 
MoE also has applications in computer vision \cite{NEURIPS2021_48237d9f_scaling_vision_sparse_moe,Zhang_2023_ICCV_Robust_mox_convolutional,pmlr-v202-chowdhury23a-patch-level-routing-in-moe}. However, there is few dedicated exploration of MoE in visual tracking. MoETrack \cite{tang2024revisiting_moetrack} employs multiple prediction heads as experts in the field of RGB-T tracking, which is rather far-fetched and deviates from the general practice of embedding MoE within Transformer layers. In the field of RGB-E tracking, eMoETracker \cite{chen2024emoetracker} inserts multiple experts between Transformer layers, which more closely resembles an adapter architecture \cite{houlsby2019parameter_adapter} and  contrasts with  MoE implementations where experts are embedded within Transformer layers.

\subsection{Parameter-efficient Fine-tuning}


Parameter-efficient fine-tuning aims to reduce computational costs and retain the general knowledge of the pretrained model by freezing most parameters and fine-tuning only a small subset of parameters. Common parameter-efficient fine-tuning approaches include adapter-based methods \cite{houlsby2019parameter_adapter,gao2024clip-adapter,pfeiffer2020AdapterHub_mad_x}, prompt-based methods \cite{jia2022visualPromptTuning,zhou2022learning_to_prompt,wang2022learningtopromptforcontinual,li2021prefix} and low-rank adaptation methods \cite{hu_lora,tian2024hydralora}. In visual tracking, HIPTrack \cite{Cai_2024_CVPR_HIPTrack} incorporates historical target features as prompts, while LoRAT \cite{LoRAT} employs low-rank adaptation for efficient training. 
And our method maintains flexibility in diverse relation modeling while reducing trainable parameters by fine-tuning TMoE.

\section{Method}
\label{sec:method}


\subsection{Overall Architecture}

As illustrated in Figure \ref{fig:pipeline}, we propose SPMTrack, a novel tracker that integrates multi-frame spatio-temporal context and employs TMoE for parameter-efficient fine-tuning.
The overall architecture follows the one-stream paradigm, consisting of a Feature Extraction Network based on a Vision Transformer \cite{dosovitskiy2020image} as backbone, and a prediction head.  

\noindent\textbf{Feature Extraction Network.}  The feature extraction network is denoted as $\Phi$. The main body of the feature extraction network is a Transformer encoder that utilizes TMoEBlock as its primary component, which is designed based on TMoE. The architecture of TMoEBlock is illustrated in the right panel of Figure \ref{fig:pipeline}, and details  will be introduced in Section \ref{section3.2}.
The encoder simultaneously performs feature extraction and relation modeling for input images. We extend the input from one single template $\bm{Z} \in \mathbb{R}^{H_z \times W_z \times 3}$ 
that is commonly used in one-stream trackers to the video sequence $\{\bm{Z}_1,...,\bm{Z}_N | \bm{Z}_i \in \mathbb{R}^{H_z \times W_z \times 3}\}$ as reference frames. The reference frames contain the first frame template and the selected tracked historical frames, with a total of $N$ frames.
The feature extraction network uses patch embedding to partition each input image into patches of size $M \times M$, map all patches into tokens by a convolutional layer, and obtain the reference token sequence of each image $\bm{T}_i \in \mathbb{R}^{N_T \times d}$, where $d$ denotes the hidden dimension of Transformer encoder and $N_T=\frac{H_zW_z}{M^2}$ represents the number of tokens corresponding to each image. Similarly, for the search region $\bm{S}\in \mathbb{R}^{H_s \times W_s \times 3}$, we can also obtain the token sequence $\bm{X} \in \mathbb{R}^{N_X \times d}$. 
All tokens are added with positional embedding and a learnable token type embedding \cite{LoRAT}. Inspired by  \cite{lin2022swintrack,ODTrack}, we additionally introduce a target state token $\bm{H} \in \mathbb{R}^{1\times d}$ that is learnable as well and propagates over time. All tokens are concatenated and fed into the Transformer encoder. The process can be formulated as: 

\vspace{-2ex}
\begin{equation}\label{eq_1}\small
\begin{aligned}
\bm{T}_{i}^j = & \begin{cases}
\bm{T}_{i}^j + \bm{PE}_{j} + \bm{TE}_o, \ if \ \text{token} \ (i,j) \ \text{in bbox}    \\
\bm{T}_{i}^j + \bm{PE}_{j} + \bm{TE}_b, \ otherwise \\
\end{cases} \\
&\bm{X}^j = \bm{X}^j + \bm{PE}_j + \bm{TE}_S \\
& \bm{I} = \mathrm{Concat}(\bm{H},\bm{T}_1,...,\bm{T}_N,\bm{X})
\end{aligned}
\end{equation}

\noindent where $\bm{T}_i^j$ represents the $j^{th}$ token in $\bm{T}_i$, $\bm{PE}_j$ represents positional embedding of the $j^{th}$ token, similarly for $\bm{X}^j$. $\bm{TE}_o$, $\bm{TE}_b$ and $\bm{TE}_S$ represents the learnable type embeddings of the foreground region tokens, background region tokens in reference frames and the search region tokens, respectively. $\bm{I}$ denotes the input to the Transformer encoder. 

Subsequently, the Transformer encoder processes all tokens and produces an output $\bm{O} \in \mathbb{R}^{(1+ N \times N_T + N_X) \times d}$. The output tokens $\bm{X}' \in \mathbb{R}^{N_X \times d}$  corresponding to search region and the  output $\bm{H}' \in \mathbb{R}^{1 \times d}$ from the target state token are fed into the prediction head for prediction. 
$\bm{H}'$ will be added with the input target state token in the next tracking frame, which can be formulated as:

\vspace{-1ex}
\begin{equation}\label{eq_2}\small
\bm{O} = \Phi(\bm{I}), \ \ [\bm{H}',\bm{T}_1',...,\bm{T}_N',\bm{X}'] = \mathrm{Split}(\bm{O})
\end{equation}

\noindent\textbf{Prediction Head.} 
The input to the prediction head consists of the output target state token $\bm{H}'$ and the output search region tokens $\bm{X}'$ from the feature extraction network. Initially, the prediction head performs matrix multiplication between $\bm{H}'$ and $\bm{X}'$ to obtain the weight $\bm{U} \in \mathbb{R}^{N_X \times 1}$. As the target state token propagates over time, $\bm{U}$ can re-weight and adjust current search region tokens based on the historical state of the target, thereby introducing more spatio-temporal tracking context. After element-wise multiplication of $\bm{X}'$ with $\bm{U}$, the resulting data is reshaped into a feature map $\bm{F}\in \mathbb{R}^{\frac{H_s}{M}\times \frac{W_s}{M} \times d}$, which is then processed by a decoupled double-$\mathrm{MLP}$ head to perform target center classification and box regression, respectively. For fair comparison, the double-$\mathrm{MLP}$ head is identical to \cite{LoRAT}. 

\subsection{The Design of TMoEBlock}\label{section3.2}
TMoEBlock serves as the primary component of the Transformer encoder in the feature extraction network. As illustrated in the right of Figure \ref{fig:pipeline}, TMoEBlock is built based on the standard vision Transformer block. However, unlike the standard ViT block, TMoEBlock applies TMoE module to both the multi-head self-attention (MSA) layer and the feed-forward network (FFN) layer, obtaining MoE-based MSA (M$^2$SA) layer and MoE-based FFN (MFFN) layer. The finer-grained application of TMoE is capable to achieve more diverse combinations for various relation modeling. The processing of TMoEBlock can be formulated as:

\begin{equation}\label{eq_3}\small
\begin{aligned}
\bm{O}'_{l} = & \mathrm{M^2SA}(\mathrm{LN}(\bm{O}_{l-1})) + \bm{O}_{l-1}, \quad l=1,...,L \\
\bm{O}_{l} &=  \mathrm{MFFN}(\mathrm{LN}(\bm{O}'_{l})) + \bm{O}'_{l}, \quad l=1,...,L
\end{aligned}
\end{equation}


\noindent where $L$ denotes the total number of TMoEBlocks, $\bm{O}_l$ represents the output of the $l^{th}$ block in Transformer encoder.  When $l=L$ , $\bm{O}_l$ is the output $\bm{O}$ of the feature extraction network; when $l=0$, it is the overall input $\bm{I}$. $\mathrm{LN}$ stands for the layer normalization operation.

\noindent\textbf{MoE-based Multi-head Self-Attention (M$^2$SA).} 
M$^2$SA not only replaces the standard three linear projection layers $\bm{Q}$, $\bm{K}$, $\bm{V}$ in standard attention for the query, key, and value with TMoE modules but also replaces the output projection layer with TMoE module.


\noindent\textbf{MoE-based Feed-Forward Network (MFFN).} 
Conventional feed-forward network consists of two linear layer and activation function.
MFFN replaces both linear layers in FFN with TMoE modules, while maintaining all other architectural configurations unchanged.

\begin{figure}[t]
\centering
\includegraphics[scale=0.5]{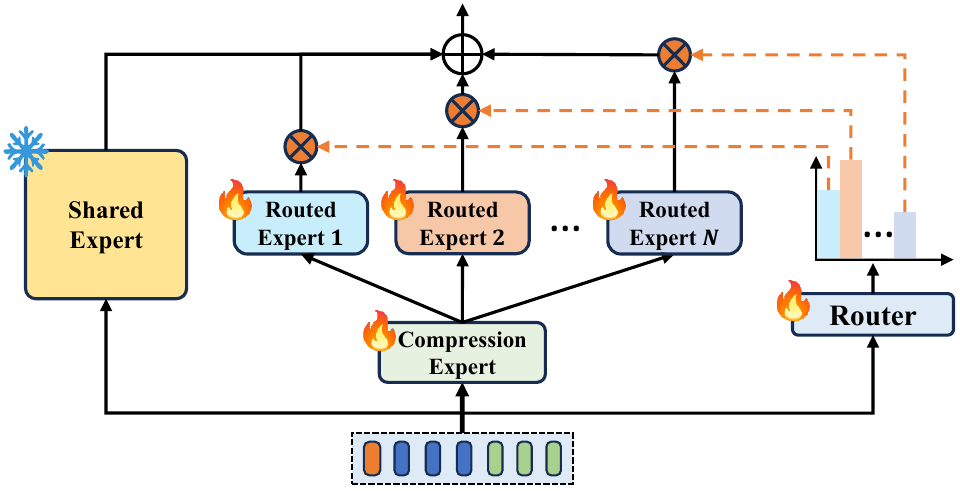}
\vspace{-1ex}
\caption{The structure of TMoE. The symbols maintain the same meaning with Figure \ref{fig:pipeline}.}
\vspace{-1ex}
\label{fig:tmoe}
\end{figure}

\subsection{Mixture of Experts for Visual Tracking}\label{section3.3}
In this section, we introduce TMoE, a mixture of experts module tailored for visual tracking task. As illustrated in Figure \ref{fig:tmoe}, TMoE consists of one shared expert, one router, one compression expert and $N_e$ routed experts. In large language models, mixture of experts are commonly applied to replace FFN layers, and the experts comprising MoE are typically the same structure as FFN. However, TMoE is used to replace the linear layers in both self-attention and FFN, and the shared expert, the compression expert and the routed experts are all implemented as linear layers. 


In TMoE, the shared expert directly copies weights from the corresponding  linear layers in pretrained model and remains frozen during training. 
The rationale is that the original weights have been well-trained during the pre-training stage, and the shared expert needs to retain more general knowledge. In contrast, the compression experts, routed experts, and the router are all trainable. 

Given a TMoE module replacing a linear layer with input dimension $d$ and output dimension $D$, for each token $\bm{x} \in \mathbb{R}^{d}$, as shown in Figure \ref{fig:tmoe}, TMoE first computes the weights $\bm{W} \in \mathbb{R}^{N_e}$ of all routed experts based on the input $\bm{x}$, the weights are normalized by $\mathrm{Softmax}$.
After that, TMoE calculates the output $\bm{y}^s \in \mathbb{R}^{D}$ of the shared expert and the outputs $\bm{y}^e_i  \in \mathbb{R}^{D}$ of all routed experts. In particular, instead of adopting the same structure as the shared expert, inspired by \cite{tian2024hydralora},
we first employ a compression expert with a $d\times r$ linear transformation to reduce the input dimension from $d$ to $r$ and obtain the compressed result $\bm{y}^c \in \mathbb{R}^{r}$, where $r \ll d$ and $r \ll D$. Subsequently, all routed experts use the compressed input $\bm{y}^c$ to produce their respective outputs $\bm{y}^e_i \in \mathbb{R}^D$. The process can be formally described as:

\begin{equation}\label{eq_tmoe_train}\small
\begin{aligned}
\bm{W}  = & \mathrm{Softmax}(\mathrm{Router}(\bm{x})), \quad \bm{y}^s = \bm{E}^S(\bm{x}) \\
\bm{y}^{c} = & \bm{E}^{C}(\bm{x}), \quad
 \bm{y}^e_i = \bm{E}^{R}_i(\bm{y}^c) \quad i = 1,...,N_e \\
\end{aligned}
\end{equation}

\noindent where $\mathrm{Router}$ is also a linear layer that maps the input $\bm{x}$ to weights $\bm{W}$, and $\bm{E}^S$, $\bm{E}^C$, and $\bm{E}^R_i$ denote the shared expert, compress expert, and $i^{th}$ routed expert, respectively.


After obtaining the outputs of all experts, we compute a weighted sum of the outputs $\bm{y}^e_i$ from all routed expert based on $\bm{W}$ to obtain $\bm{y}^e \in \mathbb{R}^{D}$, which means that the routed experts are densely activated in TMoE. Subsequently, $\bm{y}^e$ is added with the shared expert output $\bm{y}^s$ to get the final output $\bm{y} \in \mathbb{R}^{D}$. The process can be formulated as follows:

\begin{equation}\label{eq_tmoe_train_2}\small
\begin{aligned}
 \bm{y}^e & = \sum_{i=1}^{N_e} \bm{W}_i\bm{y}^e_i, \qquad \bm{y} =  \bm{y}^e + \bm{y}^s \\
\end{aligned}
\end{equation}

\subsection{Loss Function}
The outputs of the decoupled double-$\mathrm{MLP}$ prediction head are supervised using binary cross-entropy loss for target center classification and Generalized IoU loss \cite{rezatofighi2019generalized} for bounding box regression, respectively.
Both loss terms are assigned with the weighting coefficients set to 1.

\section{Experiments}

\subsection{Implementation Details}

\textbf{Model settings.} 
We provide three versions of SPMTrack: SPMTrack-B, SPMTrack-L, and SPMTrack-G, utilizing ViT-B \cite{dosovitskiy2020image}, ViT-L \cite{dosovitskiy2020image}, and ViT-G \cite{oquab2024dinov2} as backbones respectively. All versions maintain a consistent patch size $M$ of 14 and adopt the pretrained weights from DINOv2 \cite{oquab2024dinov2}. 
The resolution of input images are the same  across all versions, with reference frames of 196$\times$196 and search region of 378$\times$378. A crop factor of 2 is applied to reference frames and 5 to search region. We set the number of reference frames $N$ to 3 for spatio-temporal modeling. Each TMoE module incorporates 4 routed experts, where the input dimension of the routed experts $r$ is set to 64. 
Other configurations and model information are shown in Table \ref{table:model_settings}, where $L$ is the number of TMoEBlocks, $d$ denotes hidden dimension and $N_h$ is the number of attention heads. 

\begin{table}[!h]\small
    \centering
    \caption{ Model configurations, parameter counts and computational amount across SPMTrack variants of different scales.}
    \setlength{\tabcolsep}{0.5mm}
    \begin{tabular}{c|cccc}
    \Xhline{2pt}
         & \textbf{SPMTrack-B} & \textbf{SPMTrack-L} & \textbf{SPMTrack-G}   \\
        \Xhline{1pt}
        \textbf{Backbone} & $\left[\begin{array}{l}L=12 \\  d=768 \\ N_{h}=12
 \end{array}\right]$ & $\left[\begin{array}{l}L=24 \\  d=1024 \\ N_{h}=16
 \end{array}\right]$ & $\left[\begin{array}{l}L=40 \\  d=1536 \\ N_{h}=24 
 \end{array}\right]$\\
        \Xhline{0.5pt}
        \textbf{\#Params (M)} & 115.3 & 379.6 & 1339.5 \\
        \Xhline{0.5pt}
        \textbf{\#Trainable} & \multirow{2}{*}{29.2}  & \multirow{2}{*}{75.9} & \multirow{2}{*}{204.0} \\
        \textbf{Params (M)} & & & \\
        
    \Xhline{2pt}
    \end{tabular}
    \label{table:model_settings}
\end{table}

\noindent\textbf{Datasets.}
Following previous works \cite{LoRAT, Cai_2024_CVPR_HIPTrack, Wei_2023_CVPR_autoregressive, ye_2022_joint}, we use LaSOT \cite{fan2019lasot}, TrackingNet \cite{wang2020tracking}, GOT-10K \cite{huang2019got}, and COCO \cite{lin2014microsoft} for training. When evaluating on GOT-10K \emph{test} set, we strictly follow  the protocol by training exclusively on GOT-10K \emph{training} set. When evaluating on other datasets, we jointly train on the \emph{training} sets of all four datasets, randomly sampling from each dataset with equal probability.  For video datasets, we sample 5 frames from one video with random frame intervals ranging from 1 to 200. The sampled frames are arranged either in forward or reverse order with equal probability, where the first 3 frames serve as reference frames and the remaining 2 as search frames. For COCO, we replicate each image 5 times to maintain consistency with the video frame sampling strategy.

\noindent\textbf{Training and Optimization.}
Our method is implemented based on PyTorch 2.3.1 and trained on 8 NVIDIA A100 GPUs. We maintain a global batch size of 128 during training. When training exclusively on GOT-10K, we train for 100 epochs. When joint training on multiple datasets, we extend to 170 epochs. In each epoch, we sample 131,072 sequences. We employ AdamW \cite{DBLP:conf/iclr/LoshchilovH19_AdamW} 
 as optimizer with learning rate of $10^{-4}$ and weight decay of 0.1. The learning rate scheduler warms up over the initial 2 epochs from $10^{-7}$ to $10^{-4}$ and then follows a cosine schedule to $10^{-6}$.

\noindent\textbf{Inference.} 
We use a total of 3 reference frames, including the annotated first frame template. When current search frame index $t \le 3$, we use all available tracked frames as reference frames, if there are fewer than three frames, we repeat the template as necessary to reach the required number. When $t > 3$, we select two tracked frames as reference frames at indexes $\lfloor\frac{t}{3}\rfloor $ and $\lfloor\frac{2t}{3}\rfloor $. Following previous trackers \cite{ye_2022_joint, ODTrack, Cai_2023_ICCV_ROMTrack}, we apply a Hanning window penalty to the output response map of the classification head.  

\begin{table*}[!th]\small
    \centering
    \caption{State-of-the-art comparison on LaSOT, GOT-10k and TrackingNet. `*' denotes for trackers trained only with GOT-10k \emph{train} split. 
 The best three results are highlighted in \textbf{\textcolor{red}{red}}, \textbf{\textcolor{blue}{blue}} and \textbf{bold}, respectively.}
\setlength{\tabcolsep}{1.5mm}
    \label{whole_comparison}
    \begin{tabular}{c|c|cccccccccccc}
        \Xhline{2pt}
        Method & Source & \multicolumn{3}{c}{LaSOT} & \multicolumn{3}{c}{GOT-10k*} & \multicolumn{3}{c}{TrackingNet} \\
        \cmidrule(lr){3-5}\cmidrule(lr){6-8}\cmidrule(lr){9-11} &
         & AUC(\%) & $P_{Norm}$(\%) & $P$(\%) & AO(\%) & $SR_{0.5}$(\%) & $SR_{0.75}$(\%) & AUC(\%) & $P_{Norm}$(\%) & $P$(\%)\\
        \Xcline{1-1}{0.4pt}
        \Xhline{1pt}
        \textbf{SPMTrack-B} & \textbf{Ours} & 74.9 & 84.0 & 81.7 & 76.5 & 85.9 & 76.3 & \textbf{86.1} & 90.2 & 85.6 \\
        \textbf{SPMTrack-L} & \textbf{Ours} & \textbf{\textcolor{blue}{76.8}} & \textbf{\textcolor{blue}{85.9}} & \textbf{\textcolor{blue}{84.0}} & \textbf{\textcolor{blue}{80.0}} & \textbf{\textcolor{red}{89.4}} & \textbf{79.9} & \textbf{\textcolor{blue}{86.9}} & \textbf{\textcolor{blue}{91.0}} & \textbf{\textcolor{blue}{87.2}}\\
        \textbf{SPMTrack-G} & \textbf{Ours} & \textbf{\textcolor{red}{77.4}} & \textbf{\textcolor{red}{86.6}} & \textbf{\textcolor{red}{85.0}} & \textbf{\textcolor{red}{81.0}} & \textbf{\textcolor{blue}{89.2}} & \textbf{\textcolor{red}{82.3}} & \textbf{\textcolor{red}{87.3}} & \textbf{\textcolor{red}{91.4}} & \textbf{\textcolor{red}{88.1}} \\

        \hline
        LoRAT-B$_{378}$ \cite{LoRAT} & ECCV24 & 72.9 & 81.9 & 79.1 & 73.7 & 82.6 & 72.9 & 84.2 & 88.4 & 83.0 \\
        LoRAT-L$_{378}$ \cite{LoRAT} & ECCV24 & 75.1 & 84.1 & 82.0 & 77.5 & 86.2 & 78.1 & 85.6 & 89.7 & 85.4\\
        LoRAT-G$_{378}$ \cite{LoRAT} & ECCV24 & \textbf{76.2} & \textbf{85.3} & \textbf{83.5} & 78.9 & 87.8 & \textbf{\textcolor{blue}{80.7}} & 86.0 & 90.2 & 86.1\\
        AQATrack$_{384}$ \cite{Xie_2024_CVPR_AQATrack} & CVPR24 & 72.7 &82.9 & 80.2 & 76.0 & 85.2 & 74.9 & 84.8 & 89.3 & 84.3\\
        ARTrackV2-B$_{384}$ \cite{Bai_2024_CVPR_ARTrackV2} & CVPR24 & 73.0 &
82.0 & 79.6 & 77.5 & 86.0 & 75.5 & 85.7 & 89.8 & 85.5 \\
        ARTrackV2-L$_{384}$ \cite{Bai_2024_CVPR_ARTrackV2} & CVPR24 & 73.6 & 82.8 & 81.1 & \textbf{79.5} & 87.8 & 79.6 & 86.1 & \textbf{90.4} & \textbf{86.2}\\
        HIPTrack \cite{Cai_2024_CVPR_HIPTrack} & CVPR24 & 72.7 & 82.9 & 79.5 & 77.4 & \textbf{88.0}& 74.5 & 84.5 & 89.1 & 83.8 \\
        ODTrack-B \cite{ODTrack} & AAAI24 & 73.2 & 83.2 & 80.6  & 77.0 & 87.9 & 75.1 & 85.1 & 90.1 & 84.9 \\
        F-BDMTrack-384 \cite{Yang_2023_ICCV_FBDMTrack} & ICCV23 & 72.0 & 81.5 & 77.7 & 75.4 & 84.3 & 72.9 & 84.5 & 89.0 & 84.0\\
        ROMTrack-384 \cite{Cai_2023_ICCV_ROMTrack} & ICCV23 & 71.4 & 81.4 & 78.2 & 74.2 & 84.3 & 72.4 & 84.1 & 89.0 & 83.7 \\
        ARTrack$_{384}$  \cite{Wei_2023_CVPR_autoregressive} & CVPR23 & 72.6 & 81.7 & 79.1 & 75.5 & 84.3 & 74.3 & 85.1 & 89.1 & 84.8 \\
        SeqTrack-B$_{384}$   \cite{Chen_2023_CVPR_seqtrack} & CVPR23 & 71.5 & 81.1 & 77.8 & 74.5 & 84.3 & 71.4 & 83.9 & 88.8 & 83.6 \\
        GRM   \cite{Gao_2023_CVPR_GRM} & CVPR23 & 69.9 & 79.3 & 75.8 & 73.4 & 82.9 & 70.4 & 84.0 & 88.7 & 83.3 \\
        OSTrack$_{384}$  \cite{ye_2022_joint} & ECCV22 & 71.1 & 81.1 & 77.6 & 73.7 & 83.2 & 70.8 & 83.9 & 88.5 & 83.2 \\
        AiATrack  \cite{gao2022aiatrack} & ECCV22 & 69.0 & 79.4 & 73.8 & 69.6 & 80.0 & 63.2 & 82.7 & 87.8 & 80.4 \\
        \Xhline{2pt}
    \end{tabular} 
    \vspace{-2ex}
\end{table*}

\subsection{Comparisons with the State-of-the-Art}

\begin{figure}[!t]
\centering
\includegraphics[width=\linewidth]{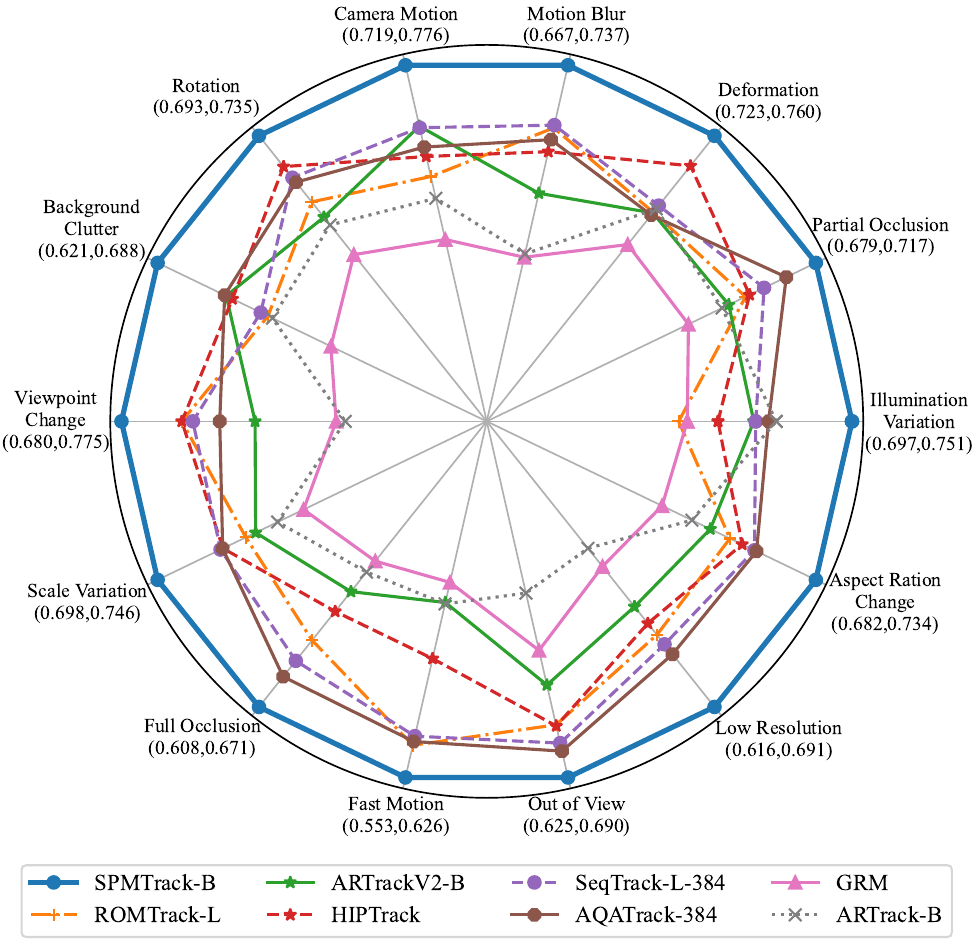}
\caption{The performance of our method compared with other state-of-the-art trackers in terms of AUC across various scenarios in the LaSOT \emph{test} split.}
\label{fig:lasotSubset}
\end{figure}

\textbf{LaSOT} \cite{fan2019lasot} is a large-scale long-term tracking dataset 
with its \emph{test} set containing 280 sequences.
As shown in Table \ref{whole_comparison}, SPMTrack-B significantly outperforms all trackers utilizing ViT-B as backbone.  Our method also substantially surpasses LoRAT \cite{LoRAT} that also employs parameter-efficient fine-tuning.  Additionally, as shown in Figure \ref{fig:lasotSubset}, we evaluate tracker performance across various challenging scenarios in LaSOT. Our method surpasses previous trackers that also incorporate spatio-temporal context \cite{Chen_2023_CVPR_seqtrack, Bai_2024_CVPR_ARTrackV2, Cai_2024_CVPR_HIPTrack,Xie_2024_CVPR_AQATrack,Wei_2023_CVPR_autoregressive}, while significantly outperforming ROMTrack \cite{Cai_2023_ICCV_ROMTrack} and GRM \cite{Gao_2023_CVPR_GRM}, which focus on optimizing relation modeling.

\noindent\textbf{GOT-10K} \cite{huang2019got} contains 9,335 sequences for training and 180 sequences for testing. We follow the protocol  that trackers are only allowed to be trained using GOT-10K \emph{train} split. Our approach significantly outperforms LoRAT \cite{LoRAT} with the same backbone. And our SPMTrack-B  rivals the performance of the current state-of-the-art methods, while SPMTrack-L substantially surpasses all existing trackers. 

\noindent\textbf{TrackingNet} \cite{muller2018trackingnet} is a large-scale visual tracking dataset with a \emph{test} set comprising 511 video sequences. Table \ref{whole_comparison} demonstrates that our method achieves state-of-the-art performance using only ViT-B as backbone, surpassing current trackers that utilize ViT-L or even ViT-G.

\noindent\textbf{TNL2K} \cite{Wang_2021_CVPR_TNL2k} comprises 700 test 
 video sequences, each accompanied by natural language descriptions. As shown in Table \ref{comparison_tnl2k}, 
 our method achieves state-of-the-art performance without utilizing natural language descriptions and surpasses existing vision-language trackers \cite{Li_2023_ICCV_citetracker, NEURIPS2022_1c8c87c3_VLT}.

\noindent\textbf{UAV123} \cite{mueller2016benchmark} is a dataset focusing on low-altitude drone aerial scenarios, consisting of 123 video sequences with an average length of 915 frames per sequence. 
Our method achieves state-of-the-art performance with comparable model sizes, demonstrating the capability of our approach in aerial scenes and small object tracking.

\noindent\textbf{NfS} \cite{Galoogahi_2017_ICCV_nfs} consists of 100 video sequences totaling 380,000 frames. Following previous works, we conduct evaluations on the 30 FPS version of the dataset. Our method rivals current state-of-the-art trackers in NfS. One possible reason for not achieving further improvement is that our method does not incorporate consecutive  target position like ARTrack \cite{Wei_2023_CVPR_autoregressive} and HIPTrack \cite{Cai_2024_CVPR_HIPTrack}, which is  crucial for NfS.

\noindent\textbf{OTB2015} \cite{otb2015} is a classic dataset in visual tracking, consisting of 100 short-term tracking sequences that encompass 11 common challenges, such as target deformation, occlusion, and scale variation. Our method also achieves state-of-the-art performance on OTB2015.

\begin{table}[]\small
    \centering
    \caption{The performance of our method and other state-of-the-art trackers on TNL2K. The best three results are highlighted in \textbf{\textcolor{red}{red}}, \textbf{\textcolor{blue}{blue}} and \textbf{bold}.}
    \setlength{\tabcolsep}{2mm}
    \begin{tabular}{c|ccc}
    \Xhline{2pt}
        \textbf{Method} & AUC(\%) & $P_{Norm}$(\%) & $P$(\%)  \\
        \Xhline{1pt}
        \textbf{SPMTrack-G} & \textbf{\textcolor{red}{64.7}} & \textbf{\textcolor{red}{82.6}} & \textbf{\textcolor{red}{70.6}} \\
        \textbf{SPMTrack-L} & \textbf{\textcolor{blue}{63.7}} & \textbf{\textcolor{blue}{81.5}} & \textbf{\textcolor{blue}{69.2}}\\
        \textbf{SPMTrack-B} & \textbf{\textbf{62.0}} & \textbf{\textbf{79.7}} & \textbf{\textbf{66.7}}  \\
        LoRAT-B$_{378}$ \cite{LoRAT} & 59.9 & - & 63.7 \\
        ODTrack-L \cite{ODTrack} & 61.7 & - & - \\
        ARTrackV2-L$_{384}$ \cite{Bai_2024_CVPR_ARTrackV2} & 61.6 & - & - \\
        ARTrack-L$_{384}$ \cite{Wei_2023_CVPR_autoregressive} & 60.3 & - & - \\
        RTracker-L \cite{Huang_2024_CVPR_RTracker} & 60.6 & - & 63.7 \\
        UNINEXT-H \cite{Yan_2023_CVPR_UNINEXT} & 59.3 & - & 62.8 \\
        CiteTracker \cite{Li_2023_ICCV_citetracker} & 57.7 & - & 59.6 \\
        VLT \cite{NEURIPS2022_1c8c87c3_VLT} & 53.1 & - & 53.3 \\
        
    \Xhline{2pt}
    \end{tabular}
    \label{comparison_tnl2k}
\end{table}

\begin{table}[h]\small
    \centering
    \caption{The performance of our method and other state-of-the-art trackers on UAV123, NfS and OTB2015 in terms of AUC metrics. The best three results are highlighted in \textbf{\textcolor{red}{red}}, \textbf{\textcolor{blue}{blue}} and \textbf{bold}.}
    \setlength{\tabcolsep}{3mm}
    \begin{tabular}{c|ccc}
    \Xhline{2pt}
        \textbf{Method} & \textbf{UAV123} & \textbf{NfS} & \textbf{OTB2015}\\
        \Xhline{1pt}
        \textbf{SPMTrack-B} & \textbf{\textcolor{red}{71.7}} & \textbf{67.4} & \textbf{\textcolor{red}{72.7}} \\
        HIPTrack \cite{Cai_2024_CVPR_HIPTrack} & \textbf{\textcolor{blue}{70.5}} & \textbf{\textcolor{red}{68.1}} & 71.0 \\
        ARTrackV2-B \cite{Bai_2024_CVPR_ARTrackV2} & 69.9 & \textbf{\textcolor{blue}{67.6}} & - \\
        ODTrack-L \cite{ODTrack} & - & - & \textbf{\textcolor{blue}{72.4}} \\
        ARTrack$_{384}$  \cite{Wei_2023_CVPR_autoregressive} & \textbf{70.5} & 66.8 & - \\
        SeqTrack-B$_{384}$  \cite{Chen_2023_CVPR_seqtrack} & 68.6 & 66.7 & - \\
        DropTrack  \cite{Wu_2023_CVPR_dropmae} & - & - & 69.6 \\
        MixFormer-L  \cite{Cui_2022_MixFormer} & 69.5 & - & - \\
        STMTrack \cite{fu2021stmtrack} & 64.7 & - & \textbf{71.9} \\
        
    \Xhline{2pt}
    \end{tabular}
    \label{table:uav,nfs,otb}
\end{table}

\subsection{Ablation Study}


\noindent\textbf{The Importance of Spatio-Temporal Context Modeling and TMoE.}
Table \ref{table_spatio-temporal_and_TMoE} demonstrates the significance of two core designs in our paper: spatio-temporal context modeling and TMoE. In Table \ref{table_spatio-temporal_and_TMoE}, removing TMoE indicates direct fine-tuning with LoRA \cite{hu_lora}, while removing spatio-temporal modeling indicates using only the first frame template and search region as input and eliminating the target state token.
Compared to LoRA, applying TMoE for parameter-efficient fine-tuning yields a substantial improvement of \textbf{+0.8} AUC on LaSOT, demonstrates the superiority of TMoE. And the spatio-temporal context modeling further enhances performance by an additional \textbf{+1.2} AUC. On the basis of incorporating spatio-temporal modeling, the improvement brought by TMoE is \textbf{+1.0} AUC, indicating that TMoE can help the model better capture spatio-temporal contextual information.

\begin{table}[h]\small
    \centering
    \caption{Ablation studies on spatio-temporal context modeling and TMoE. Experiments are conducted on LaSOT.}
    \setlength{\tabcolsep}{1.2mm}
    \begin{tabular}{c|cc|ccc}
    \Xhline{2pt}
        \textbf{\#} & \textbf{Spatio-Temporal} & \textbf{TMoE}  &  AUC (\%) & \textbf{$P_{Norm} (\%)$} & \textbf{$P (\%)$} \\
        \Xhline{1pt}
        \textbf{1} & \ding{56} & \ding{56} & 72.9 & 81.9 & 79.1 \\ 
        \textbf{2} & \ding{56}  & \ding{52} & 73.7 & 82.7 & 80.0 \\
        \textbf{3} & \ding{52} & \ding{56} & 73.9 & 82.8 & 80.0 \\
        \textbf{4} & \ding{52} & \ding{52} & \textbf{74.9} & \textbf{84.0} & \textbf{81.7} \\
        
    \Xhline{2pt}
    \end{tabular}
    \label{table_spatio-temporal_and_TMoE}
    \vspace{-2ex}
\end{table}

\noindent\textbf{Where to Apply TMoE.} 
Conventional MoE is typically applied to the FFN layers within Transformer blocks, while keeping the attention layers unchanged. In contrast, we apply TMoE in both the attention layers and the FFN layers.
Table \ref{table_where_apply_tmoe} demonstrates the impact of applying TMoE in different layers. 
Comparing the first, third, and fifth rows, applying TMoE to the query, key, value projections in attention layers and to FFN layers can both improve performance. Comparing the first, second and third rows, we find that applying TMoE in attention layers yields significantly greater benefits compared to applying it only in FFN layers.  Comparing the third and fourth rows, applying TMoE in the output projection matrix of the attention layer can bring a slight improvement.

\begin{table}[h]\small
    \centering
    \caption{Ablation studies on applying TMoE in different layers. All results are evaluated on LaSOT \emph{test} split. ``\ding{56}" means replacing TMoE with LoRA.}
    \setlength{\tabcolsep}{1.5mm}
    \begin{tabular}{c|ccc|ccc}
    \Xhline{2pt}
        \textbf{\#} & \textbf{QKV} & \textbf{Out Proj} & \textbf{FFN} & AUC (\%) & \textbf{$P_{Norm} (\%)$} & \textbf{$P (\%)$} \\
        \Xhline{1pt}
        \textbf{1} & \ding{56} & \ding{56} & \ding{56} & 73.9 & 82.8 & 80.0 \\
        \textbf{2} & \ding{56} & \ding{56} & \ding{52} & 74.1 & 83.1 & 80.8 \\
        \textbf{3} & \ding{52} & \ding{56} & \ding{56} & 74.5 & 83.6 & 81.2 \\
        \textbf{4} & \ding{52} & \ding{52} & \ding{56} & 74.6 & 83.8 & 81.4 \\
        \textbf{5} & \ding{52} & \ding{52} & \ding{52} & \textbf{74.9} & \textbf{84.0} & \textbf{81.7} \\
    \Xhline{2pt}
    \end{tabular}
    \label{table_where_apply_tmoe}
    \vspace{-2ex}
\end{table}

\begin{figure}[!t]
\centering
\includegraphics[width=\linewidth]{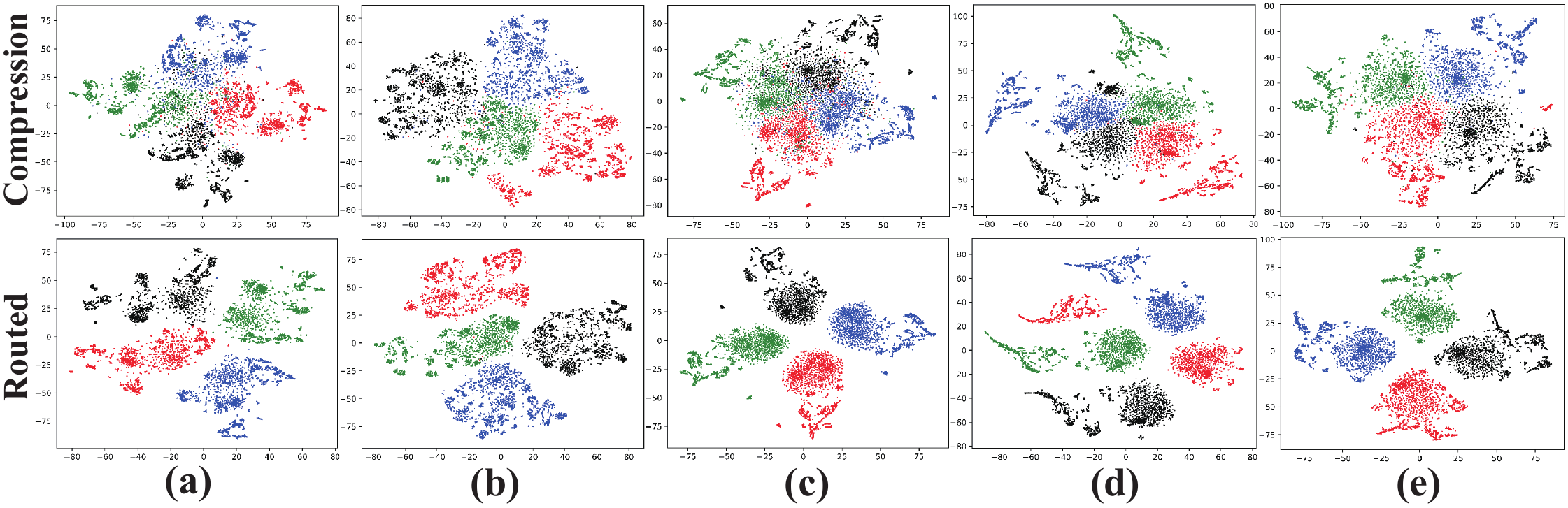}
\vspace{-3ex}
\caption{Comparison of t-SNE visualizations. Each column shows outputs from all compression experts (top) and routed experts (bottom) within a TMoE module. Different colors represent distinct experts. Zoom in for better view.}
\label{fig:tsne}
\end{figure}

\noindent\textbf{Whether to use a shared compress expert.}
In TMoE, we employ a shared compression expert to reduce the input dimension to $r$, and use multiple routed experts for subsequent processing. As shown in the first and second rows of Table \ref{table_TMoEvsMoE_and_shared_Compress_expert}, compared to providing each routed expert with a compression expert, sharing the compression expert can achieve better performance while reducing the parameter count. 
When not sharing the compression expert, we provide t-SNE visualizations of the outputs from compression experts and routed experts in five TMoE modules in Figure \ref{fig:tsne}. The outputs of different compression experts show significant overlaps, while the routed experts are more dispersed, validating the effectiveness of our approach.

\noindent\textbf{TMoE vs. Conventional MoE.}
We compare TMoE with conventional MoE on the visual tracking task. For conventional MoE, we use LoRA to fine-tune the attention layers and apply MoE in the FFN layers with five experts: one frozen shared expert that copies pretrained FFN parameters, and four trainable routed experts, all routed experts are identical to the original FFN architecture, and a router is also applied. 
As shown in the first row and third row of Table \ref{table_TMoEvsMoE_and_shared_Compress_expert}, TMoE demonstrates significant advantages over conventional MoE in both parameter efficiency and performance.  

\begin{table}[h]\small
    \centering
    \caption{Ablation studies on whether to share the compression expert and use the conventional MoE to replace TMoE. Results are evaluated on LaSOT \emph{test} split.}
    \setlength{\tabcolsep}{0.15mm}
    \begin{tabular}{c|c|ccc}
    \Xhline{2pt}
        \textbf{model variants} & \textbf{\#Params(M)} & AUC (\%) & \textbf{$P_{Norm} (\%)$} & \textbf{$P (\%)$}  \\
        \Xhline{1pt}
        SPMTrack-B &  \textbf{115.3} & \textbf{74.9} & \textbf{84.0} & \textbf{81.7} \\
        \scalebox{0.9}{\it{w/o}} \footnotesize{shared compress expert} & 131.3 & 74.8 & 83.7 & 81.5  \\
        \footnotesize{Conventional MoE} & 316.8 & 73.4 & 82.6 & 79.8  \\
    \Xhline{2pt}
    \end{tabular}
    \label{table_TMoEvsMoE_and_shared_Compress_expert}
\end{table}

\noindent\textbf{The Number of Routed Experts.}
In Table \ref{table_expert_num}, we evaluated the impact of different numbers of routed experts in TMoE. Results indicate that increasing the number of experts leads to improved performance, highlighting the potential of TMoE as a method to expand the parameter count and model scale. To balance the number of parameters and performance, we choose to use 4 experts, and whether there are separate optimal configurations for the TMoE in the attention layer and the FFN layer remains to be explored.


\noindent\textbf{The Number of Reference Frames.} In SPMTrack,  we use the template and several tracked frames as reference frames to enhance the spatio-temporal modeling ability. In Table \ref{table_ablation_reference_numbers}, we explore the impact of using different numbers
of reference frames during training and inference. During inference, if only two reference frames are used, we simply select the frame at $\lfloor\frac{t}{2}\rfloor$ as the reference frame. If four reference frames are used, we select the frames at $\lfloor\frac{t}{4}\rfloor$, $\lfloor\frac{t}{2}\rfloor$, and $\lfloor\frac{3\times t}{4}\rfloor$. As shown in Table \ref{table_ablation_reference_numbers}, better performance can be achieved when the number of reference frames used during inference remains the same as that during training. Training with more reference frames can also lead to improved performance, illustrating that our method still has substantial potential for performance enhancement. 

\begin{figure}[!t]
\centering
\includegraphics[width=\linewidth]{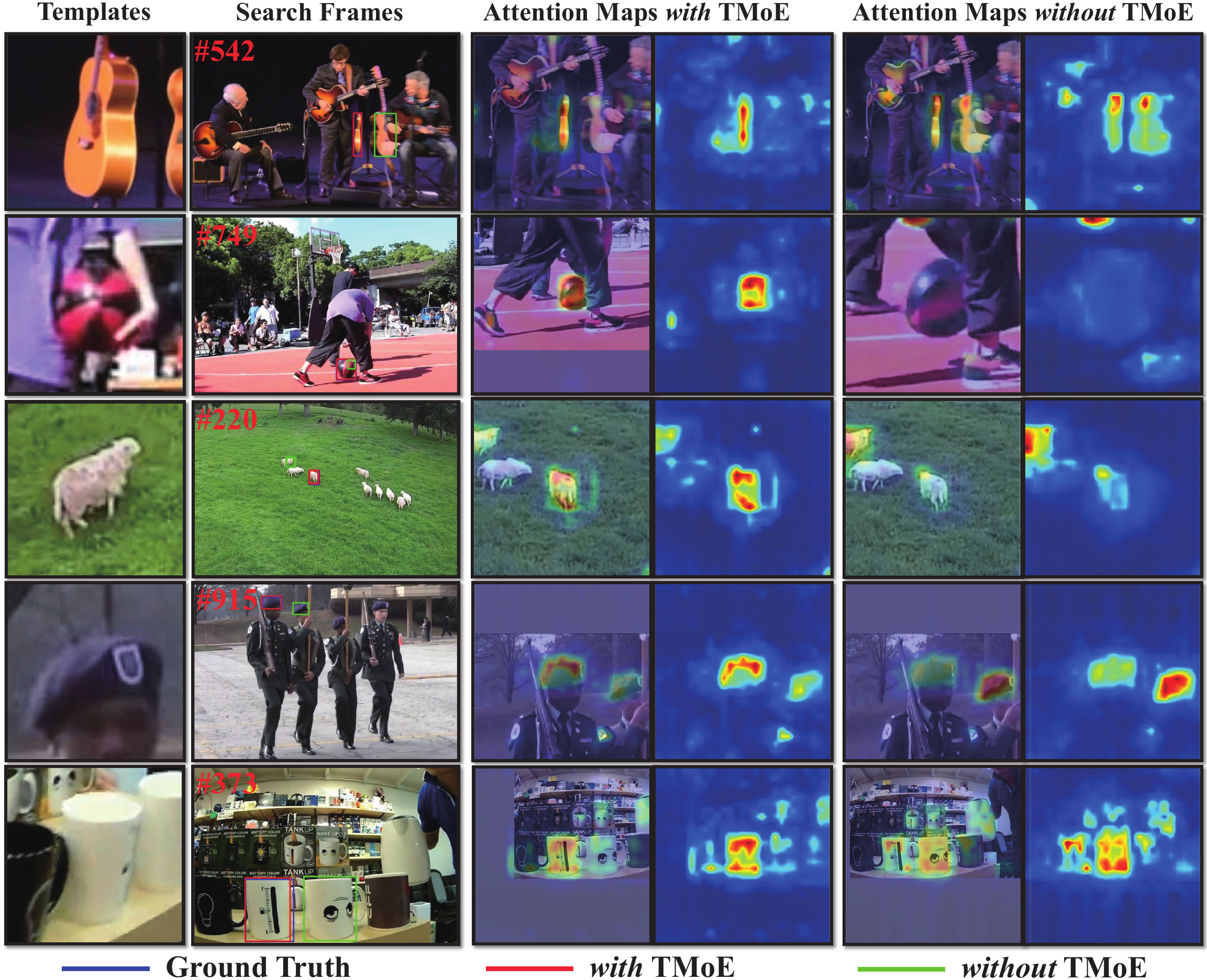}
\vspace{-2ex}
\caption{Visualization comparison of search region attention maps \emph{with} and \emph{without} TMoE. Zoom in for better view.}
\label{fig:visualize1}
\end{figure}

\begin{table}[!h]\small
    \centering
    \caption{The performance of our method on the \emph{test} split of GOT-10K
when setting different number of routed experts in TMoE.}
    \setlength{\tabcolsep}{4.5mm}
    \begin{tabular}{c|cccc}
    \Xhline{2pt}
        \textbf{Number} & \textbf{2} & \textbf{4} & \textbf{6} & \textbf{8} \\
        \Xhline{1pt}
        \textbf{AO}(\%) & 75.3 & 76.5 &  76.6 & \textbf{77.1}\\
        $\mathbf{SR_{0.5}}$(\%) & 84.4 & 85.9 &  86.3 & \textbf{86.4} \\
        $\mathbf{SR_{0.75}}$(\%) & 75.2 & 76.3 &  77.0 & \textbf{77.6} \\
        
    \Xhline{2pt}
    \end{tabular}
    \label{table_expert_num}
\end{table}

\noindent\textbf{Comparison of attention maps with and without TMoE.}
In Figure \ref{fig:visualize1}, to exclude the influence of intermediate tracked frames, we use the model in rows 1 and 2 of Table \ref{table_spatio-temporal_and_TMoE} for comparison, where spatio-temporal modeling is removed and we focus solely on the impact of TMoE. We compare the attention maps of the search region in the last block of Transformer encoder. The first column is the template, the second column displays search region, the third column is the visualization result and pure attention map, as well as the fourth column. 
Figure \ref{fig:visualize1} shows that when facing challenging scenarios such as occlusions, distractors, and viewpoint variations, TMoE effectively suppresses background regions while enhancing perception of the target boundary.


\begin{table}[]\small
    \centering
    \caption{Ablation study on the number of reference frames during training and inference. Results are evaluated based on SPMTrack-B on GOT-10K \emph{test} split.}
    \setlength{\tabcolsep}{1.7mm}
    \begin{tabular}{c|c|ccc}
    \Xhline{2pt}
        \multirow{2}{*}{\textbf{Training}} & \multirow{2}{*}{\textbf{Inference}} & \multirow{2}{*}{\textbf{AO} (\%)} & \multirow{2}{*}{\textbf{SR$\bm{_{0.5}}$}(\%)} & \multirow{2}{*}{\textbf{SR$\bm{_{0.75}}$(\%)}} \\
         &   &  &  &   \\
        \Xhline{1pt}
        \multirow{3}{*}{2} & 2  & \textbf{75.8} & \textbf{85.1} & \textbf{75.3} \\
        & 3  & 74.7 & 84.6 & 73.8 \\
        & 4  & 69.8 & 81.8 & 66.2 \\
        \Xhline{0.5pt}
        \multirow{3}{*}{3} & 2  & 73.1 & 83.9 & 72.7 \\
        & 3  & \textbf{76.5} & \textbf{85.9} & \textbf{76.3} \\
        & 4  & 72.8 &	83.4 & 71.3 \\
        \Xhline{0.5pt}
        \multirow{3}{*}{4} & 2  & 70.6 & 83.9 & 68.9 \\
        & 3  & 74.6 & 85.4 & 74.5 \\
        & 4  & \textbf{77.5} & \textbf{87.3} & \textbf{77.2} \\
    \Xhline{2pt}
    \end{tabular}
    \label{table_ablation_reference_numbers}
\end{table}
\section{Conclusion}


In this paper, we present TMoE, a mixture of experts module tailored for visual tracking, and propose SPMTrack, a novel tracker enabling spatio-temporal context modeling based on TMoE.
TMoE is applied in the linear layers in both self-attention and FFN layers, enhancing the diversity and flexibility of expert combinations to better handle various relation modeling in visual tracking.
Additionally, TMoE employs a lightweight and efficient structure and serves as a method of parameter-efficient fine-tuning, which enables us to train SPMTrack of larger scales and enables SPMTrack to achieve state-of-the-art performance with only a small subset of parameters need to be trained. Furthermore, we hope that this work will inspire more applications of mixture of experts in the field of visual tracking.

\noindent \textbf{Acknowledgements.} This paper was supported by Zhejiang Provincial Natural Science Foundation of China under Grant No.LD24F020016 and National Natural Science Foundation of China under Grant No.62176017.




\maketitlesupplementary
\appendix
\numberwithin{equation}{section}
\numberwithin{figure}{section}
\numberwithin{table}{section}
In the supplementary material, Section \ref{furtherAnalyses} presents additional ablation studies, further validating the effectiveness of other designs of SPMTrack and TMoE. In Section \ref{detailsinlasot}, we provide more comprehensive performance comparisons between our SPMTrack and other excellent trackers. The comparisons consist of success curves on both overall LaSOT \cite{fan2019lasot} \emph{test} split and all challenging scenario subsets, along with overall precision curves. And in Section \ref{qualitative}, we further showcase the comparisons of tracking results of various trackers in complex scenarios and present more qualitative analyses of our method.

\section{Further Analyses}\label{furtherAnalyses}

\subsection{Ablation Study on Target State Token}
We employ a temporally propagated target state token to store historical target states, which is utilized in the prediction head for further adjustment and refinement of search region features. In Table \ref{table_supp_target_state_token_got} and Table \ref{table_supp_target_state_token_trackingnet}, we remove the target state token and directly use the search region features from the output of feature extraction network for prediction in the prediction head. 
Table \ref{table_supp_target_state_token_got} and Table \ref{table_supp_target_state_token_trackingnet} present evaluation results on the GOT-10K \cite{huang2019got} and TrackingNet \cite{muller2018trackingnet} \emph{test} splits, respectively.
The results demonstrate that the integration of target state tokens, while introducing negligible additional parameters, improves performance on both two datasets, with a more significant improvement observed on TrackingNet.

\begin{figure*}[!t]
\centering
\includegraphics[scale=0.4]{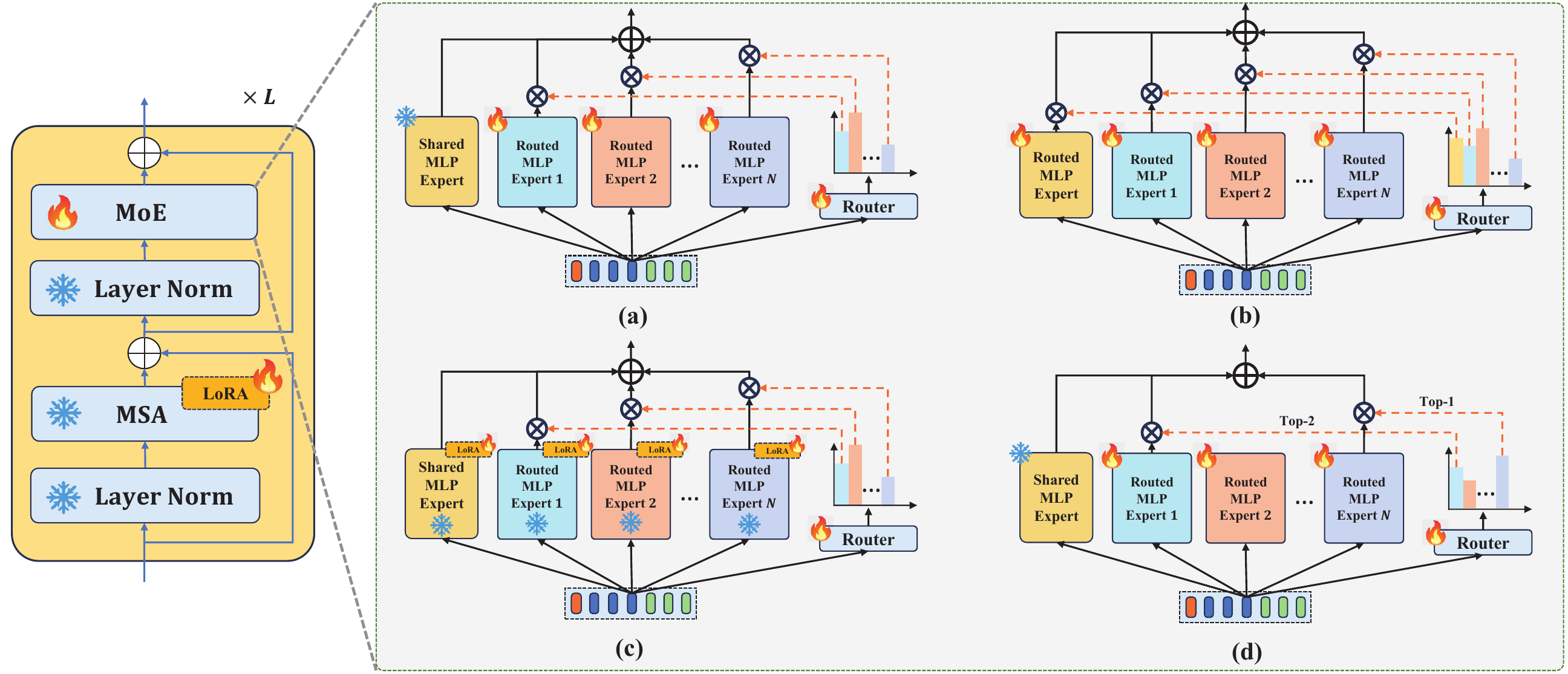}
\vspace{-3ex}
\caption{ 
Several conventional MoE architectural designs, where MoE modules exclusively replace feed-forward network (FFN) layers in the original Transformer Block, while linear layers in multi-head self-attention are fine-tuned using LoRA \cite{hu_lora}. Zoom in for better view.
}
\label{fig:conventional_moe}
\end{figure*}

\begin{table}[h]\small
    \centering
    \caption{Ablation study on whether adopt target state token. Results are evaluated on GOT-10K \cite{huang2019got} \emph{test} split.}
    \setlength{\tabcolsep}{0.5mm}
    \begin{tabular}{c|c|ccc}
    \Xhline{2pt}
        \textbf{model variants} & \textbf{\#Params(M)} & \textbf{AO} (\%) & \textbf{SR$\bm{_{0.5}}$}(\%) & \textbf{SR$\bm{_{0.75}}$(\%)}  \\
        \Xhline{1pt}
        SPMTrack-B & 115.331 & \textbf{76.5} & \textbf{85.9} & \textbf{76.3} \\
        \scalebox{0.9}{\it{w/o}} \footnotesize{target state token} & \textbf{115.329} & 76.3 & 85.6 & 75.9 \\
    \Xhline{2pt}
    \end{tabular}
    \label{table_supp_target_state_token_got}
\end{table}

\begin{table}[h]\small
    \centering
    \caption{Ablation study on whether adopt target state token. Results are evaluated on TrackingNet \cite{muller2018trackingnet} \emph{test} split.}
    \setlength{\tabcolsep}{0.5mm}
    \begin{tabular}{c|c|ccc}
    \Xhline{2pt}
        \textbf{model variants} & \textbf{\#Params(M)} & AUC (\%) & \textbf{$P_{Norm} (\%)$} & \textbf{$P (\%)$}   \\
        \Xhline{1pt}
        SPMTrack-B & 115.331 & \textbf{86.1} & \textbf{90.2} & \textbf{85.6} \\
        \scalebox{0.9}{\it{w/o}} \footnotesize{target state token} & \textbf{115.329} & 85.7 & 89.9 & 85.1 \\
    \Xhline{2pt}
    \end{tabular}
    \label{table_supp_target_state_token_trackingnet}
\end{table}

\subsection{Ablation Study on Token Type Embedding}
In SPMTrack, we add not only positional embedding but also token type embeddings to the input of the feature extraction network to further enhance the token information. 
The additional parameter count and computational overhead introduced by the token type embeddings are negligible.

In Table \ref{table_supp_token_type_embedding}, we investigate the impact of introducing different token type embeddings on the performance of SPMTrack-B.
Comparing the first row and the third row in Table \ref{table_supp_token_type_embedding}, adding three types of token type embeddings leads to a performance improvement. Comparing the first row and second row, we find that when the type embedding corresponding to the target foreground tokens is not introduced, which means all tokens in the reference frames use the same type embedding, there are no performance gains observed. The results indicate that the target foreground token type embedding is the most crucial among the three categories, as it enhances the ability of the tracker to discriminate target foreground regions, thereby improving performance.

\begin{table}[h]\small
    \centering
    \caption{Ablation study on different token type embeddings, where $\bm{TE}_o$, $\bm{TE}_b$ and $\bm{TE}_S$ represent the type embeddings of the foreground region tokens, background region tokens in reference frames and the search region tokens, respectively. }
    \setlength{\tabcolsep}{1.5mm}
    \begin{tabular}{c|ccc|ccc}
    \Xhline{2pt}
        \textbf{\#} & $\bm{TE}_o$ & $\bm{TE}_b$ & $\bm{TE}_S$  &  \textbf{AO} (\%) & \textbf{SR$\bm{_{0.5}}$}(\%) & \textbf{SR$\bm{_{0.75}}$(\%)} \\
        \Xhline{1pt}
        \textbf{1} & \ding{56} & \ding{56} & \ding{56} & 76.2 & 85.6 & 75.8 \\ 
        \textbf{2} & \ding{56}  & \ding{52} & \ding{52} &  76.2	& 85.5 & 76.0  \\
        \textbf{3} & \ding{52} & \ding{52} & \ding{52} & \textbf{76.5} & \textbf{85.9}  & \textbf{76.3} \\
        
    \Xhline{2pt}
    \end{tabular}
    \label{table_supp_token_type_embedding}
    \vspace{-2ex}
\end{table}


\subsection{Ablation Study on Different Pre-trained Models}

In this paper, we select DINOv2 \cite{oquab2024dinov2} as the pre-trained model. In Table \ref{table_supp_pretrain_model}, we investigate the impact of different pre-trained models on the final performance. All experiments are conducted based on SPMTrack-B, with pre-trained models based on the ViT-B \cite{dosovitskiy2020image} architecture. It is worth noting that for pre-trained models with an image patch size of 16, we use a reference frame size of $192\times 192$ and a search region size of $384\times 384$. 
As shown in Table \ref{table_supp_pretrain_model},
experimental results demonstrate that SPMTrack maintains consistent performance on GOT-10K regardless of the choice of pre-trained model, indicating the robustness and the generalization ability of our approach. Eventually, in order to make a fair comparison with LoRAT \cite{LoRAT} that also employs parameter-efficient fine-tuning, we still select DINOv2 \cite{oquab2024dinov2} as the pre-trained model.

\begin{table}[h]\small
    \centering
    \caption{Ablation study on different pre-trained models. Results are evaluated on GOT-10K \emph{test} split.}
    \setlength{\tabcolsep}{0.5mm}
    \begin{tabular}{c|c|ccc}
    \Xhline{2pt}
        \textbf{Pretrained Models} & \textbf{Patch Size} & \textbf{AO} (\%) & \textbf{SR$\bm{_{0.5}}$}(\%) & \textbf{SR$\bm{_{0.75}}$(\%)}  \\
        \Xhline{1pt}
        DINOv2 \cite{oquab2024dinov2} & 14 & 76.5 & 85.9 & \textbf{76.3} \\
        MAE \cite{He_2022_CVPR_mae} & 16 & 76.3 & 87.0 & 75.1 \\
        DropMAE \cite{Wu_2023_CVPR_dropmae} & 16 & \textbf{76.6} & \textbf{87.7} & 74.0 \\
    \Xhline{2pt}
    \end{tabular}
    \label{table_supp_pretrain_model}
\end{table}

\subsection{Ablation Study on MLP Prediction Head}


In SPMTrack, in order to make a fair comparison with LoRAT \cite{LoRAT}, we also adopt a fully MLP-based prediction head. 
However, this does not represent our optimal configuration. As demonstrated in Table \ref{table_supp_mlp_head}, we replace the MLP prediction head with the convolution-based prediction head used in previous trackers like OSTrack \cite{ye_2022_joint}. Although the MLP prediction head has fewer parameters, the overall model parameter difference is minimal, and the convolution-based head achieves  significantly superior performance. In contrast to LoRAT, which fails completely under parameter-efficient fine-tuning when using convolution-based heads, our method achieves even better performance with convolution-based heads, which further demonstrates the generality of our proposed TMoE for  parameter-efficient fine-tuning and also further demonstrates the potential for performance improvement in SPMTrack.


\begin{table}[h]\small
    \centering
    \caption{Ablation study on MLP-based prediction head and convolutional-based prediction head. Results are evaluated on GOT-10K \emph{test} split.}
    \setlength{\tabcolsep}{0.2mm}
    \begin{tabular}{c|c|ccc}
    \Xhline{2pt}
        \textbf{Prediction Head} & \textbf{\#Params(M)} & \textbf{AO} (\%) & \textbf{SR$\bm{_{0.5}}$}(\%) & \textbf{SR$\bm{_{0.75}}$(\%)}  \\
        \Xhline{1pt}
        MLP Head & \textbf{2.37} & 76.5 & 85.9 & 76.3 \\
        \footnotesize{OSTrack \cite{ye_2022_joint} Conv Head} & 6.47 & \textbf{76.8} &	\textbf{86.4} & \textbf{77.0} \\
    \Xhline{2pt}
    \end{tabular}
    \label{table_supp_mlp_head}
\end{table}





\begin{figure*}[!t]
\centering
{
\begin{minipage}{4.1cm}
\centering
\includegraphics[scale=0.23]{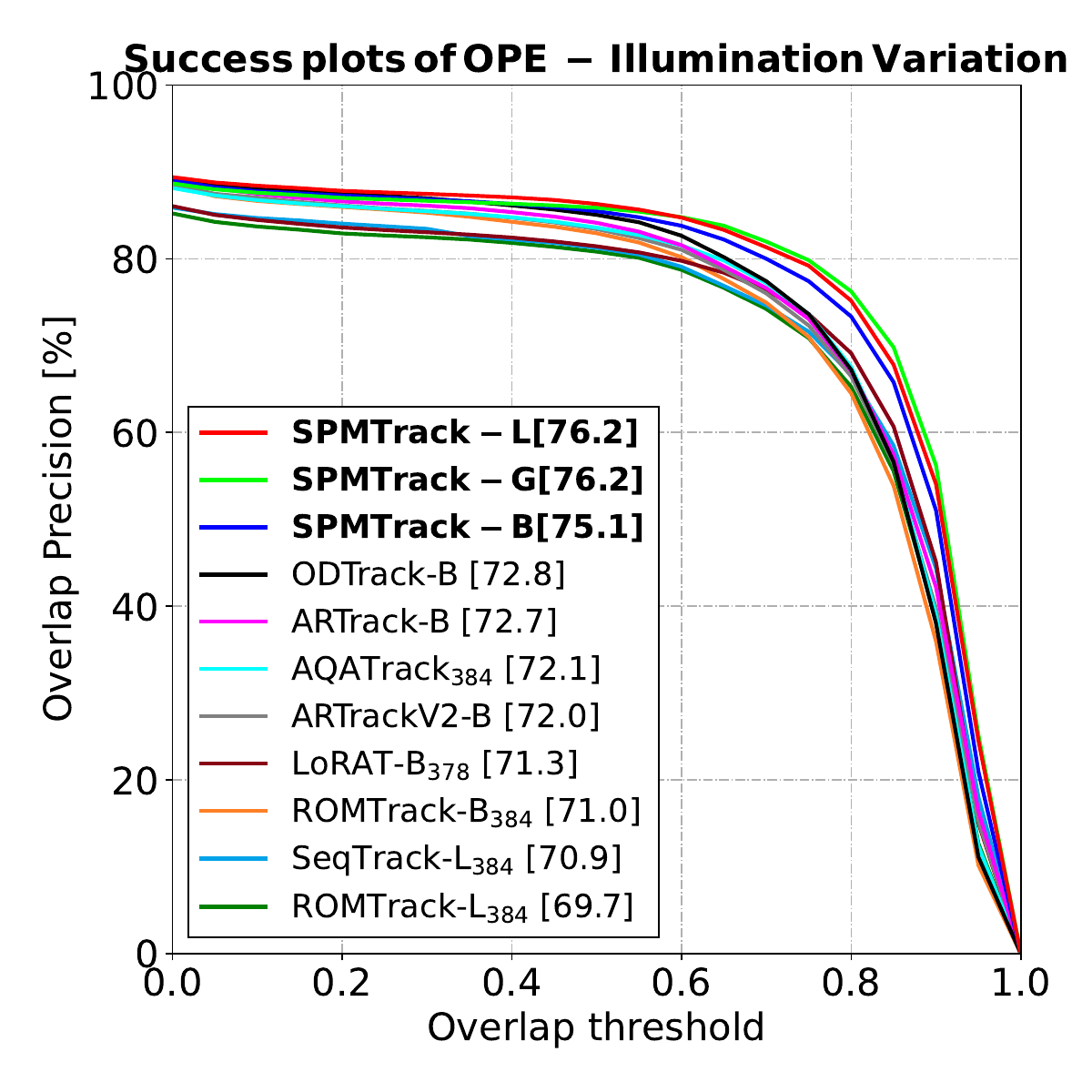} 
\end{minipage}
}
{
\begin{minipage}{4.1cm}
\centering
\includegraphics[scale=0.23]{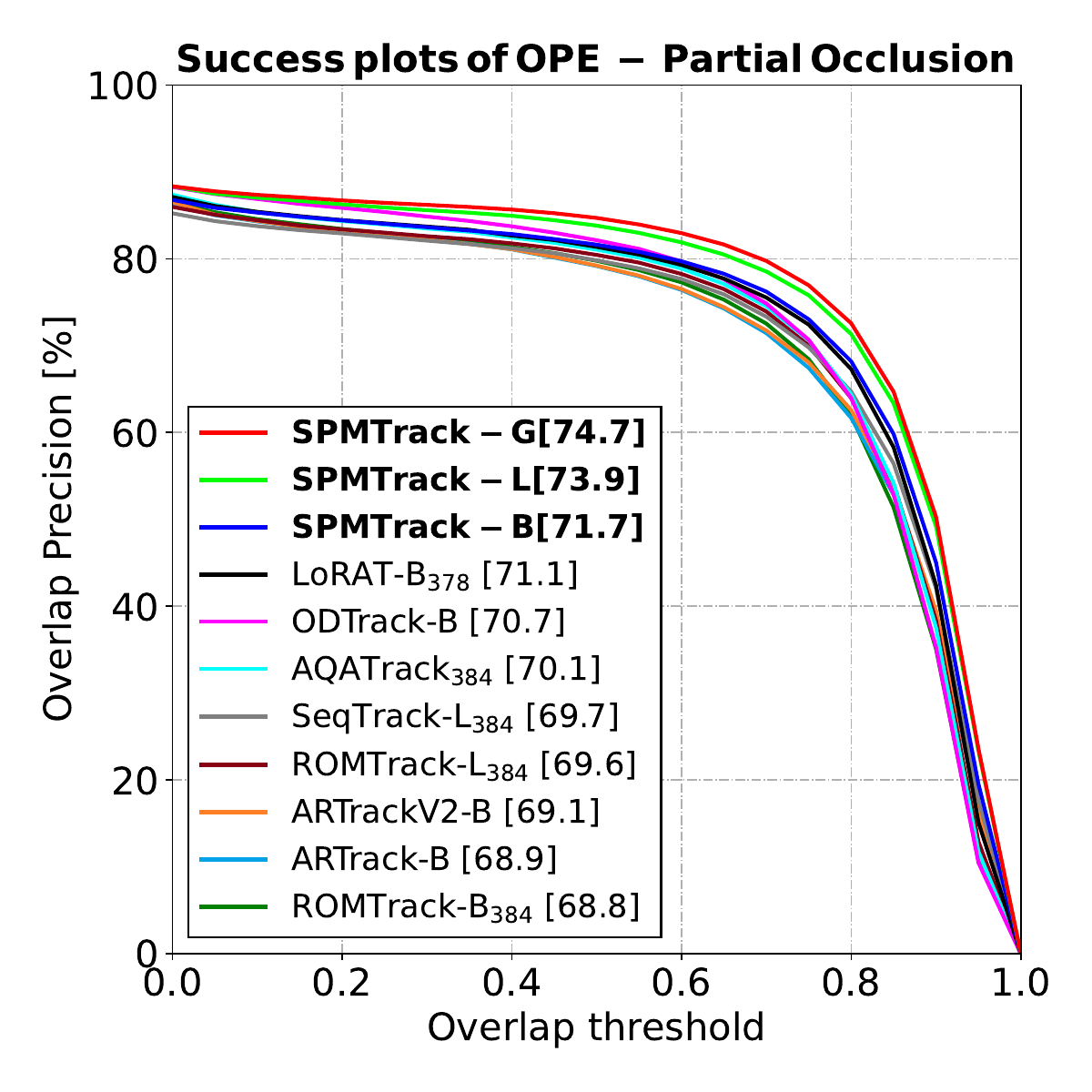}
\end{minipage}
}
{
\begin{minipage}{4.1cm}
\centering
\includegraphics[scale=0.23]{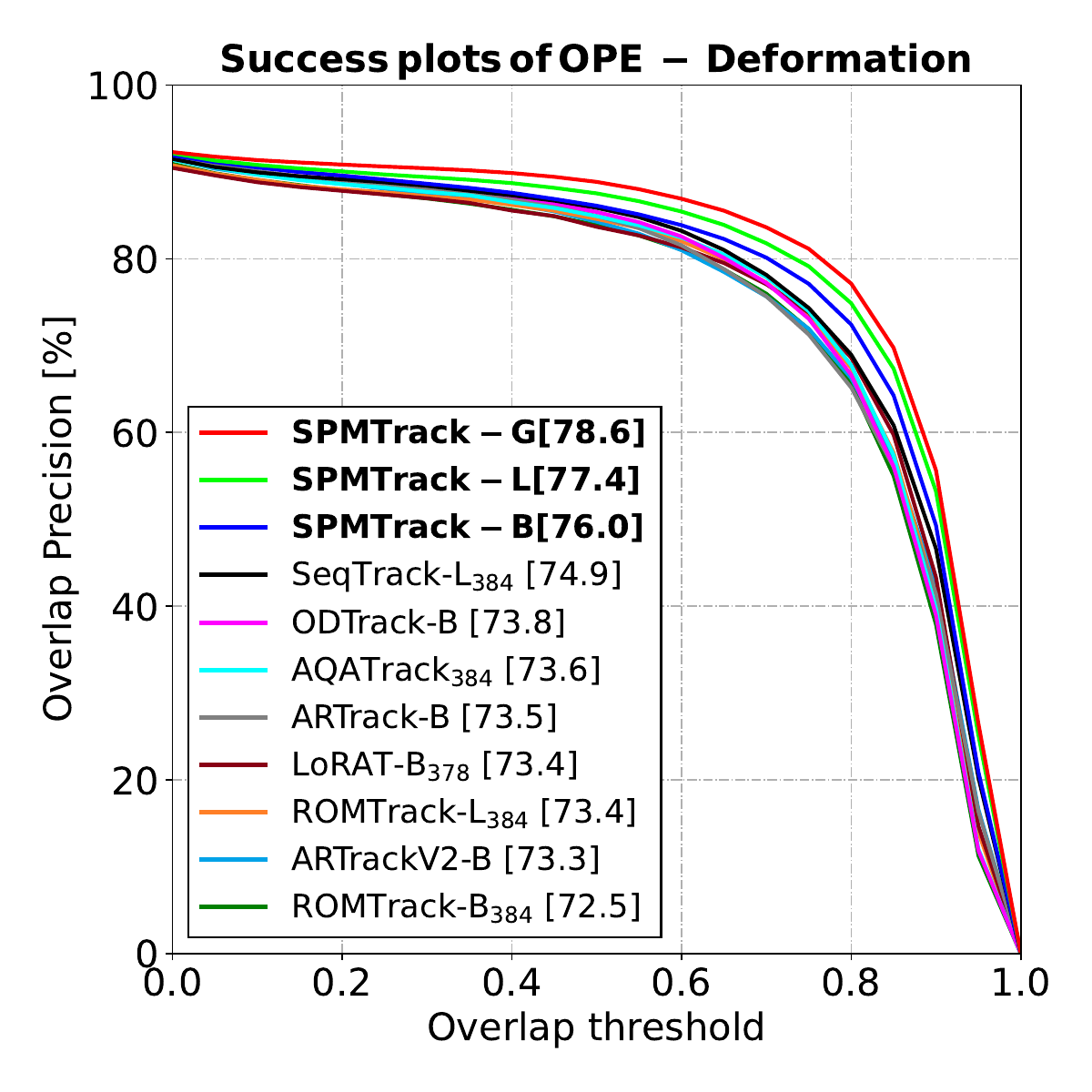}
\end{minipage}
}
{
\begin{minipage}{4.1cm}
\centering
\includegraphics[scale=0.23]{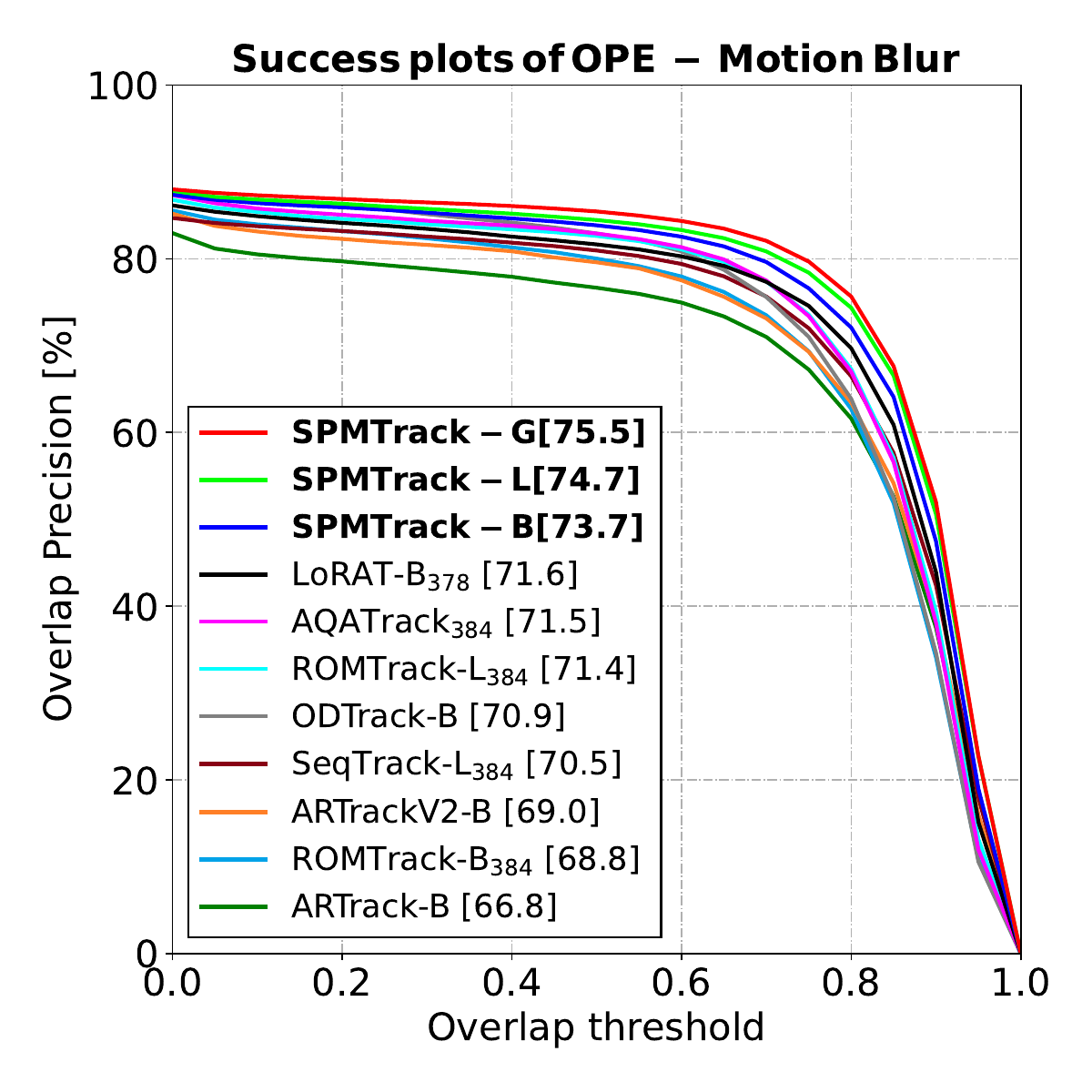}
\end{minipage}
}


{
\begin{minipage}{4.1cm}
\centering
\includegraphics[scale=0.23]{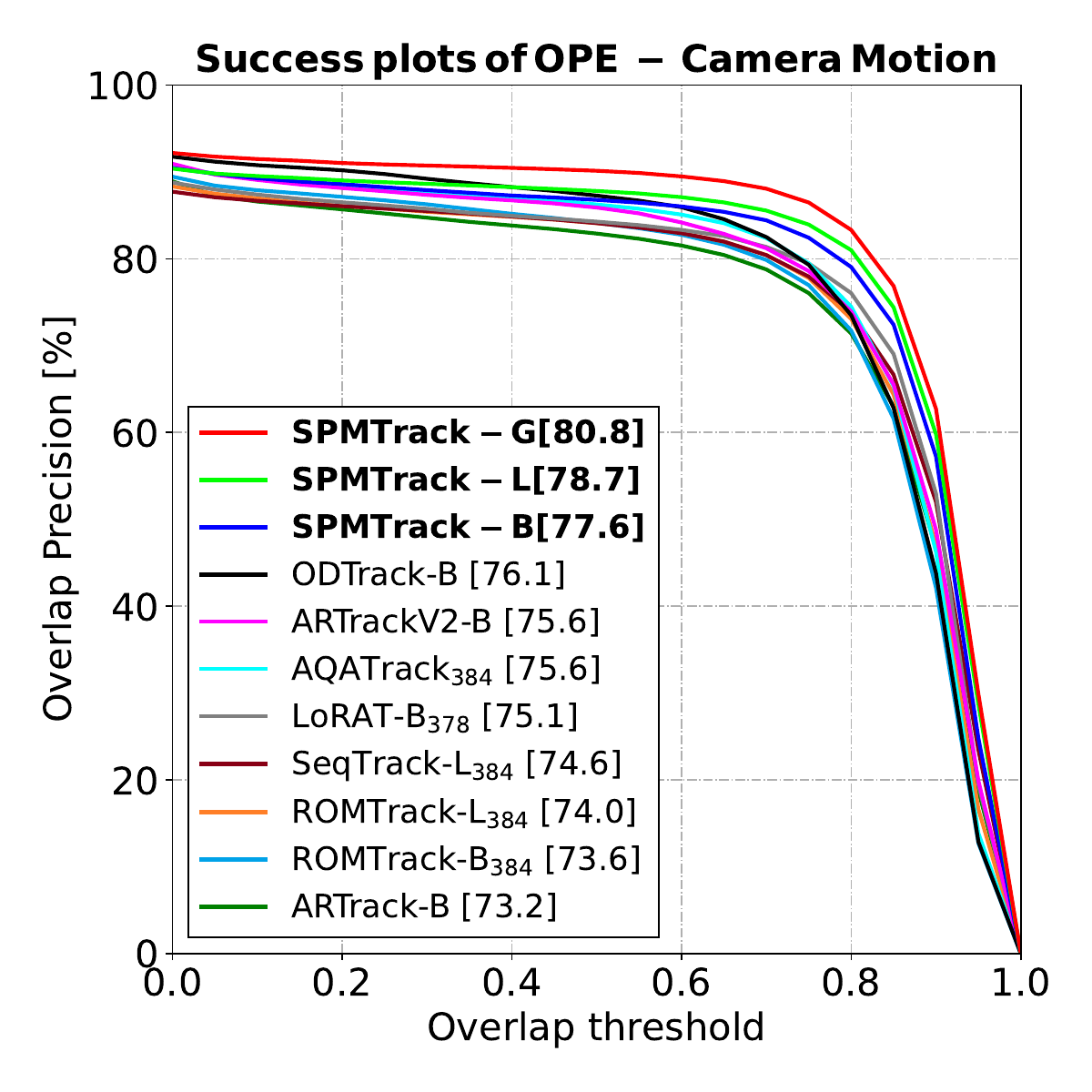}
\end{minipage}
}
{
\begin{minipage}{4.1cm}
\centering
\includegraphics[scale=0.23]{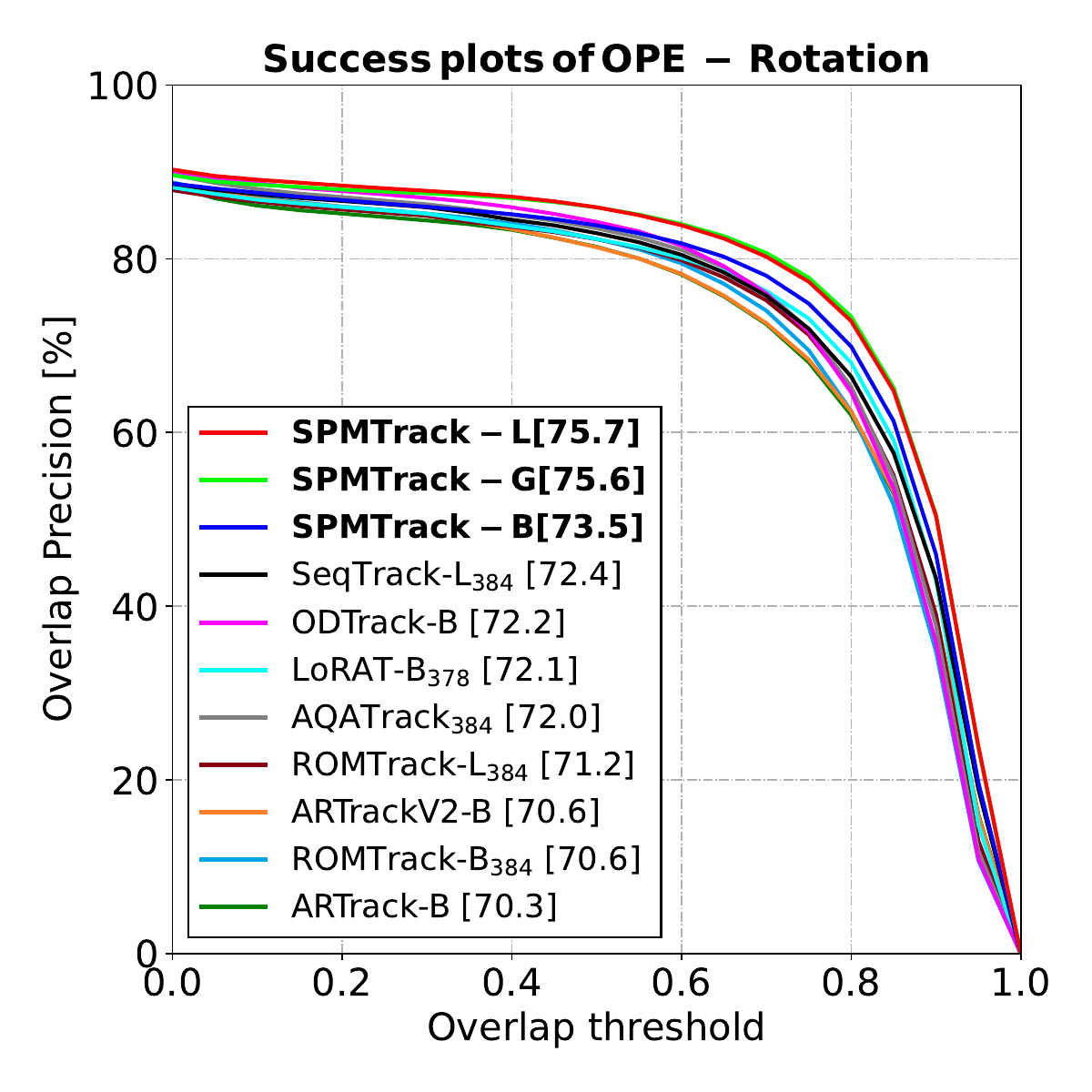}
\end{minipage}
}
{
\begin{minipage}{4.1cm}
\centering
\includegraphics[scale=0.23]{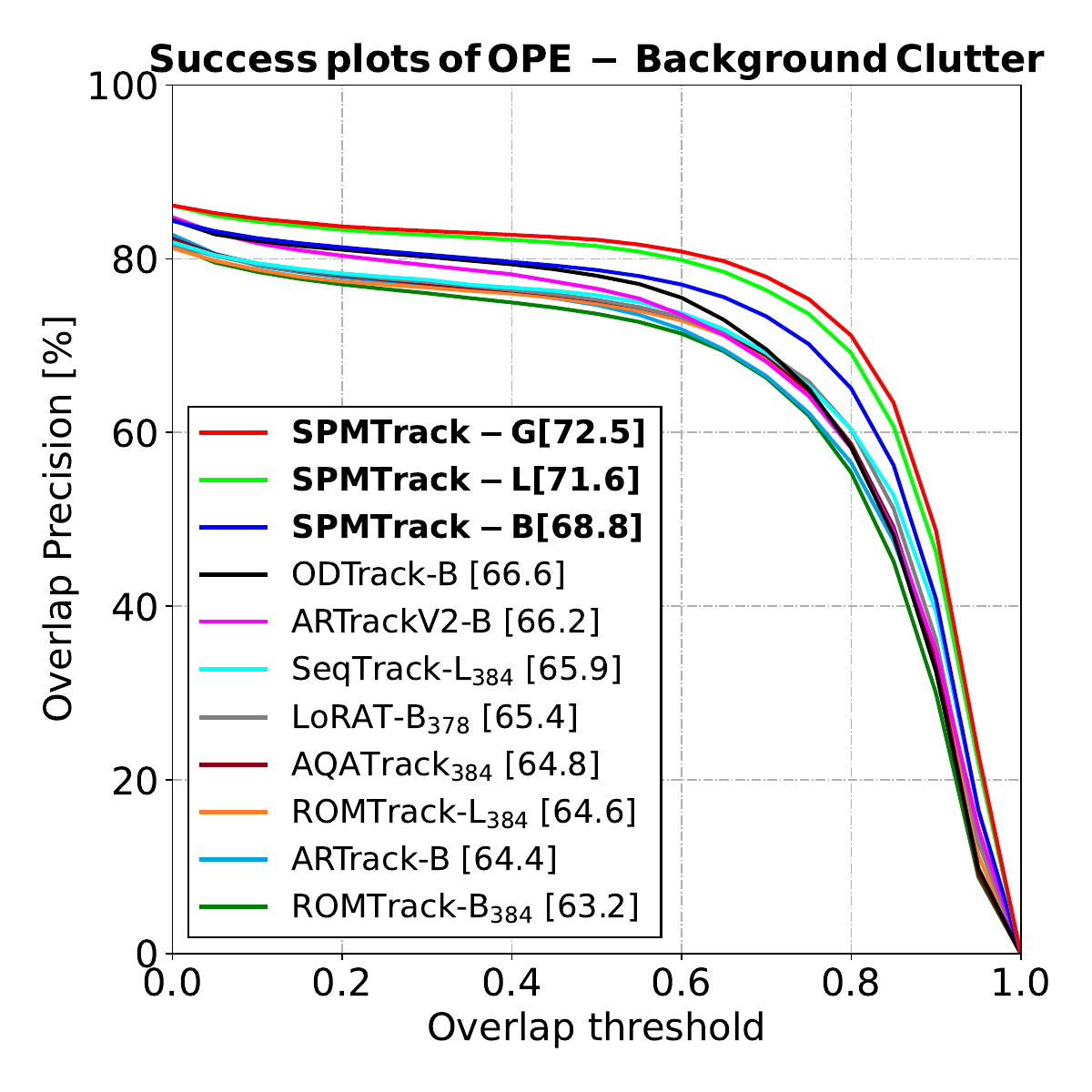}
\end{minipage}
}
{
\begin{minipage}{4.1cm}
\centering
\includegraphics[scale=0.23]{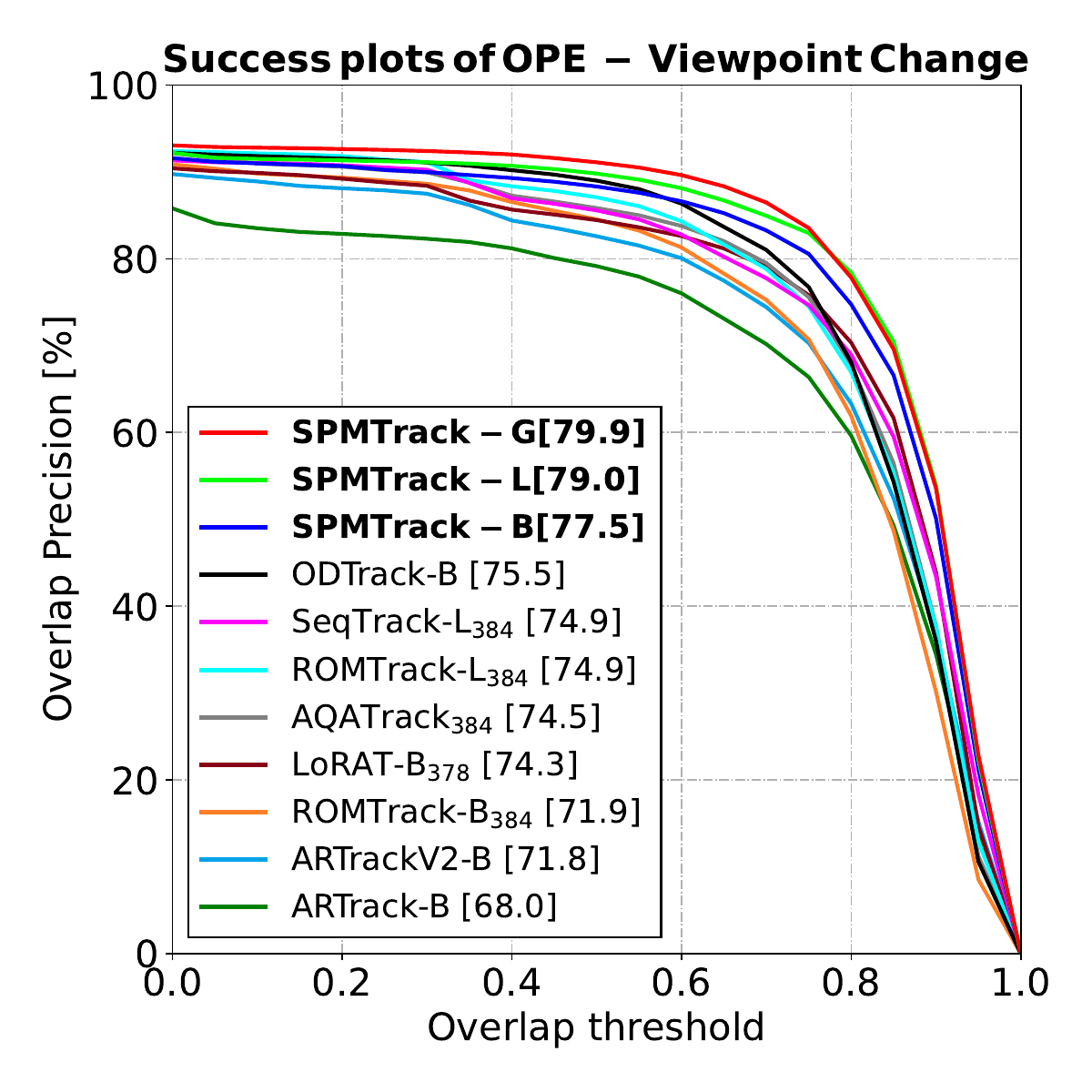}
\end{minipage}
}


{
\begin{minipage}{4.1cm}
\centering
\includegraphics[scale=0.23]{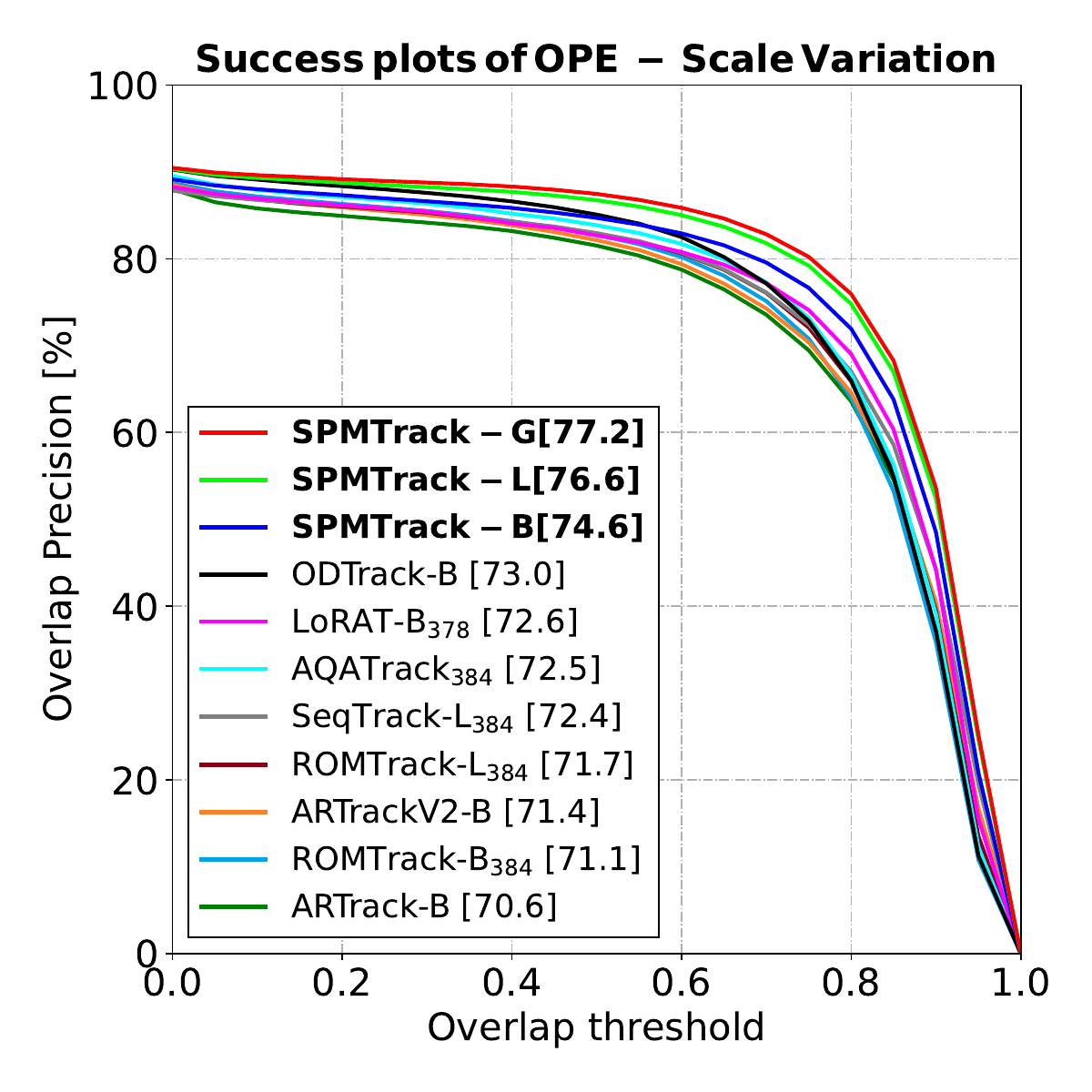}
\end{minipage}
}
{
\begin{minipage}{4.1cm}
\centering
\includegraphics[scale=0.23]{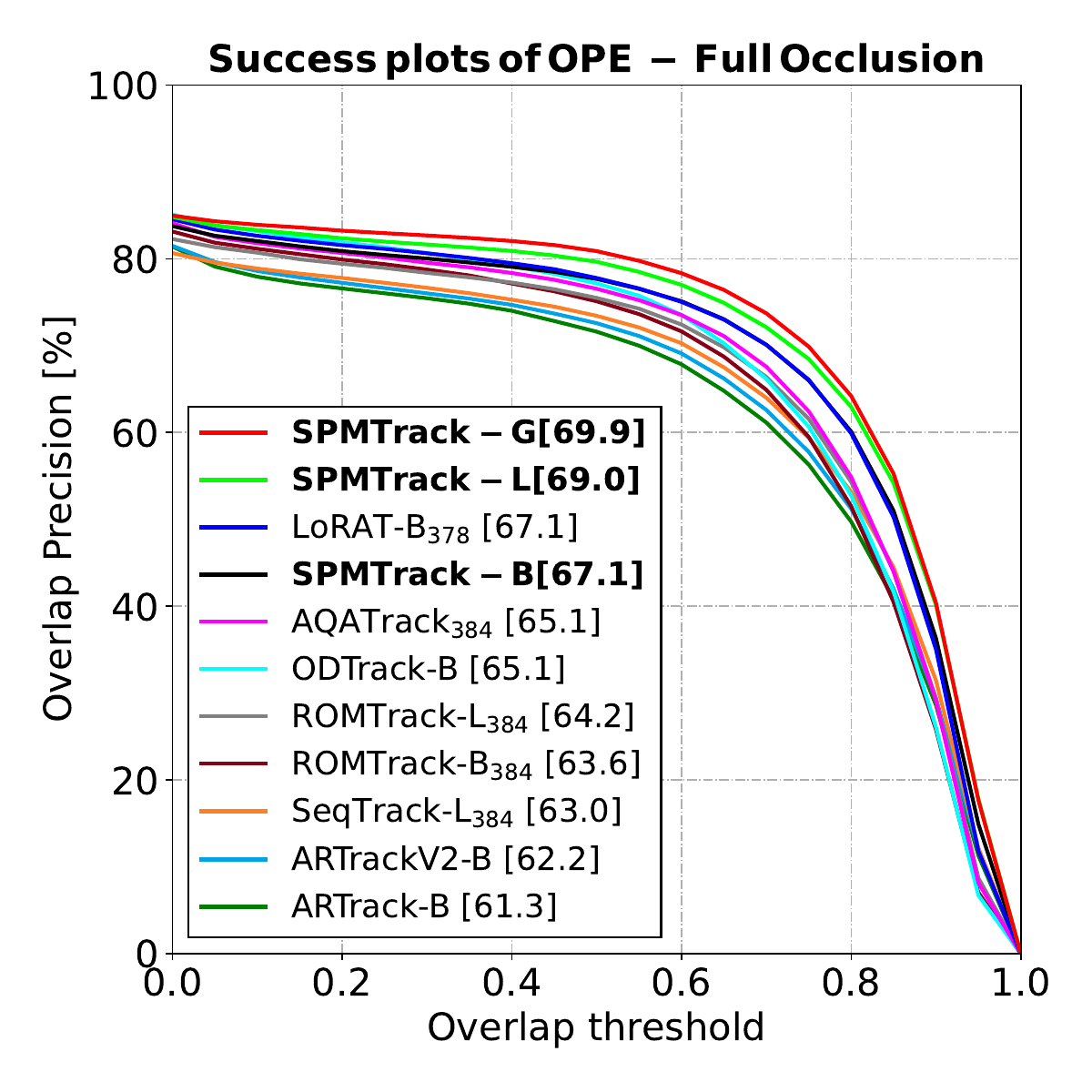}
\end{minipage}
}
{
\begin{minipage}{4.1cm}
\centering
\includegraphics[scale=0.23]{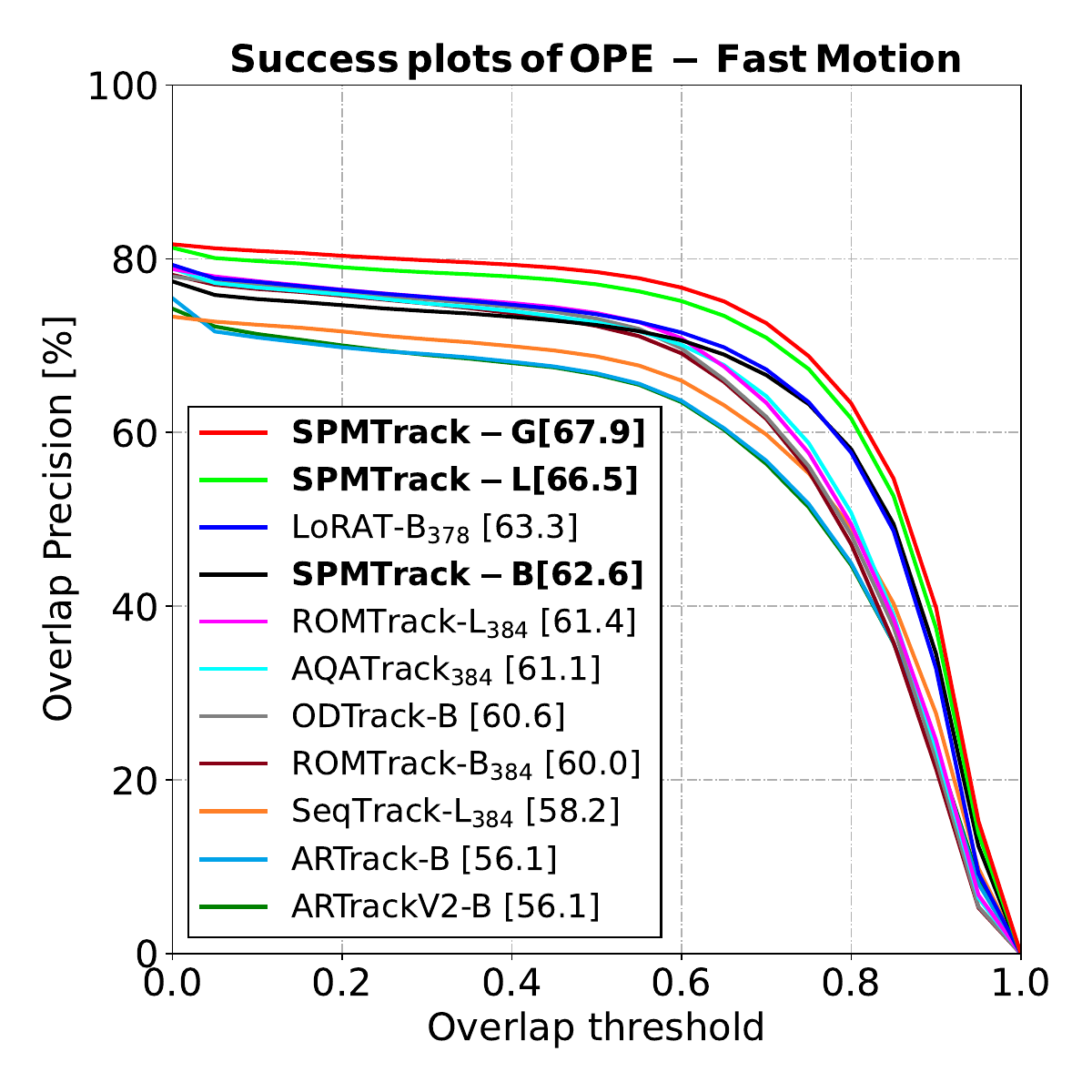}
\end{minipage}
}
{
\begin{minipage}{4.1cm}
\centering
\includegraphics[scale=0.23]{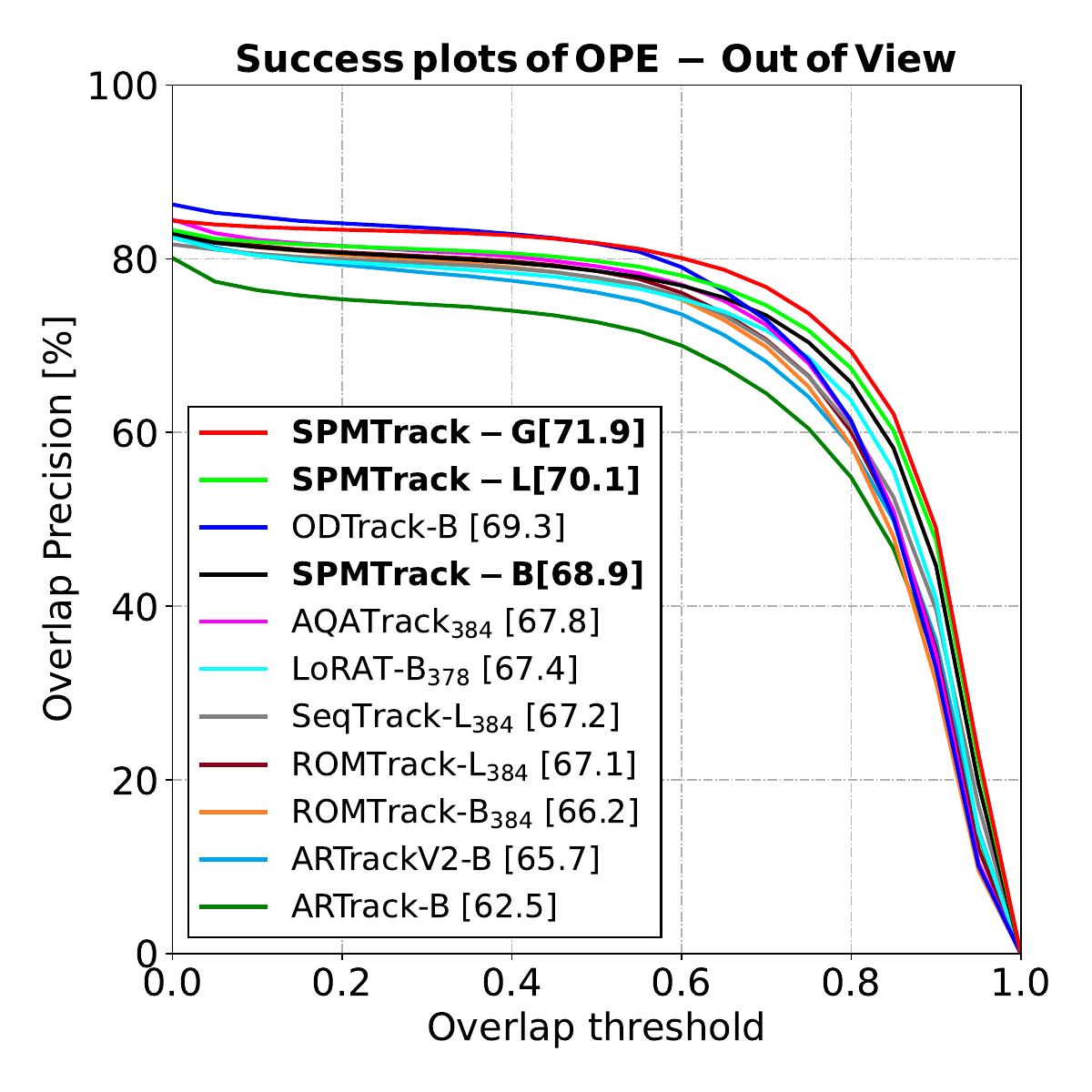}
\end{minipage}


{
\begin{minipage}{4.1cm}
\centering
\includegraphics[scale=0.23]{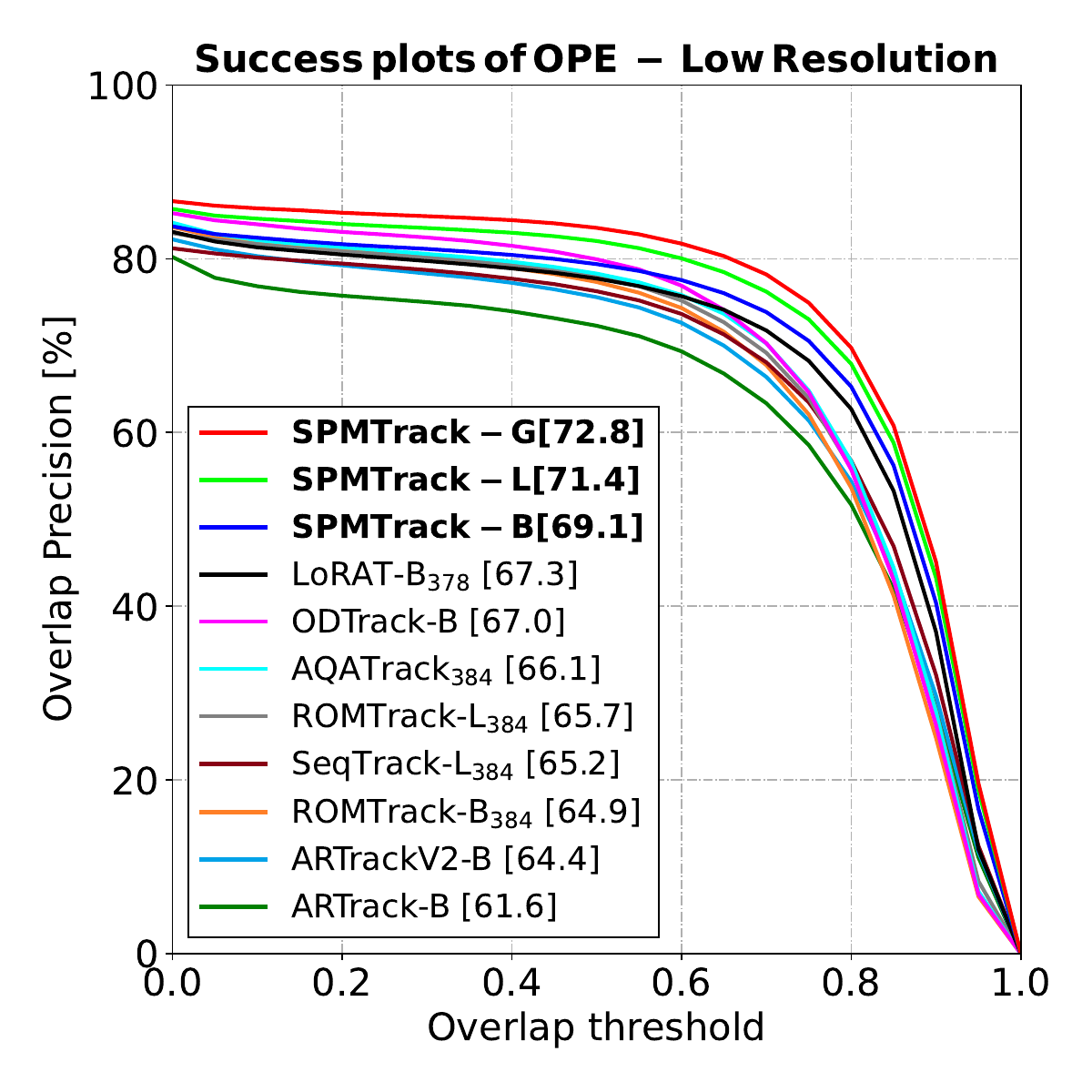}
\end{minipage}
}
{
\begin{minipage}{4.1cm}
\centering
\includegraphics[scale=0.23]{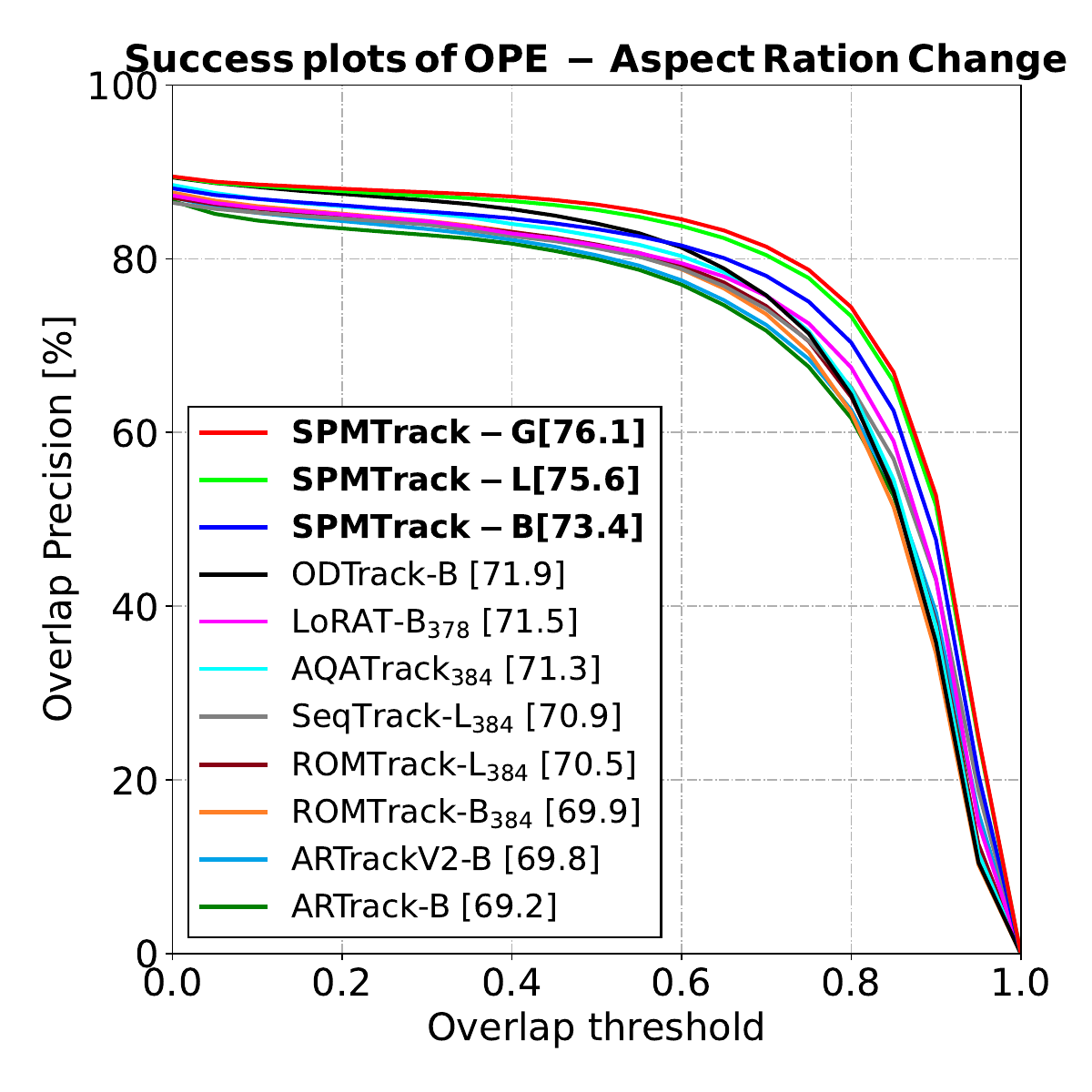}
\end{minipage}
}
{
\begin{minipage}{4.1cm}
\centering
\includegraphics[scale=0.23]{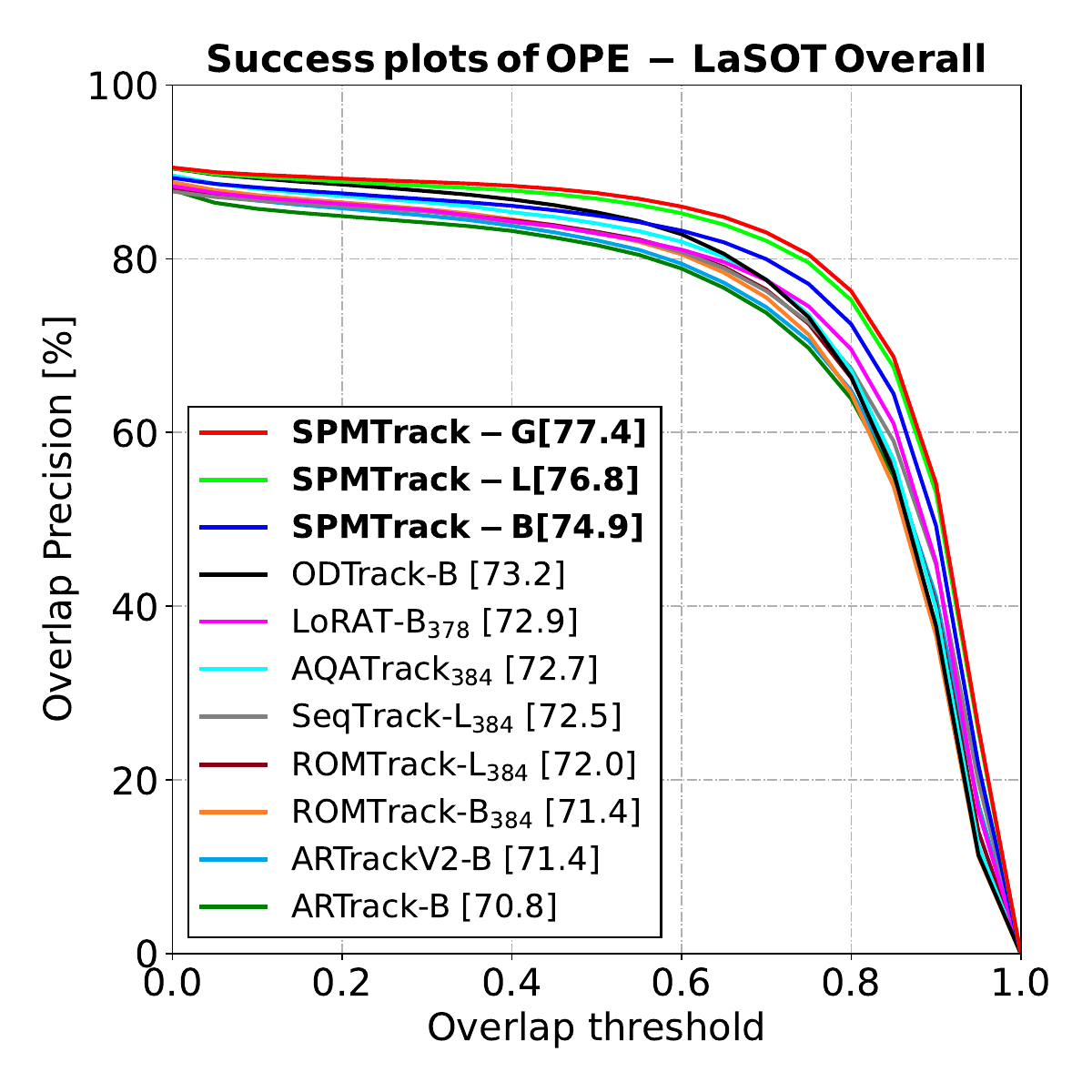}
\end{minipage}
}
{
\begin{minipage}{4.1cm}
\centering
\includegraphics[scale=0.23]{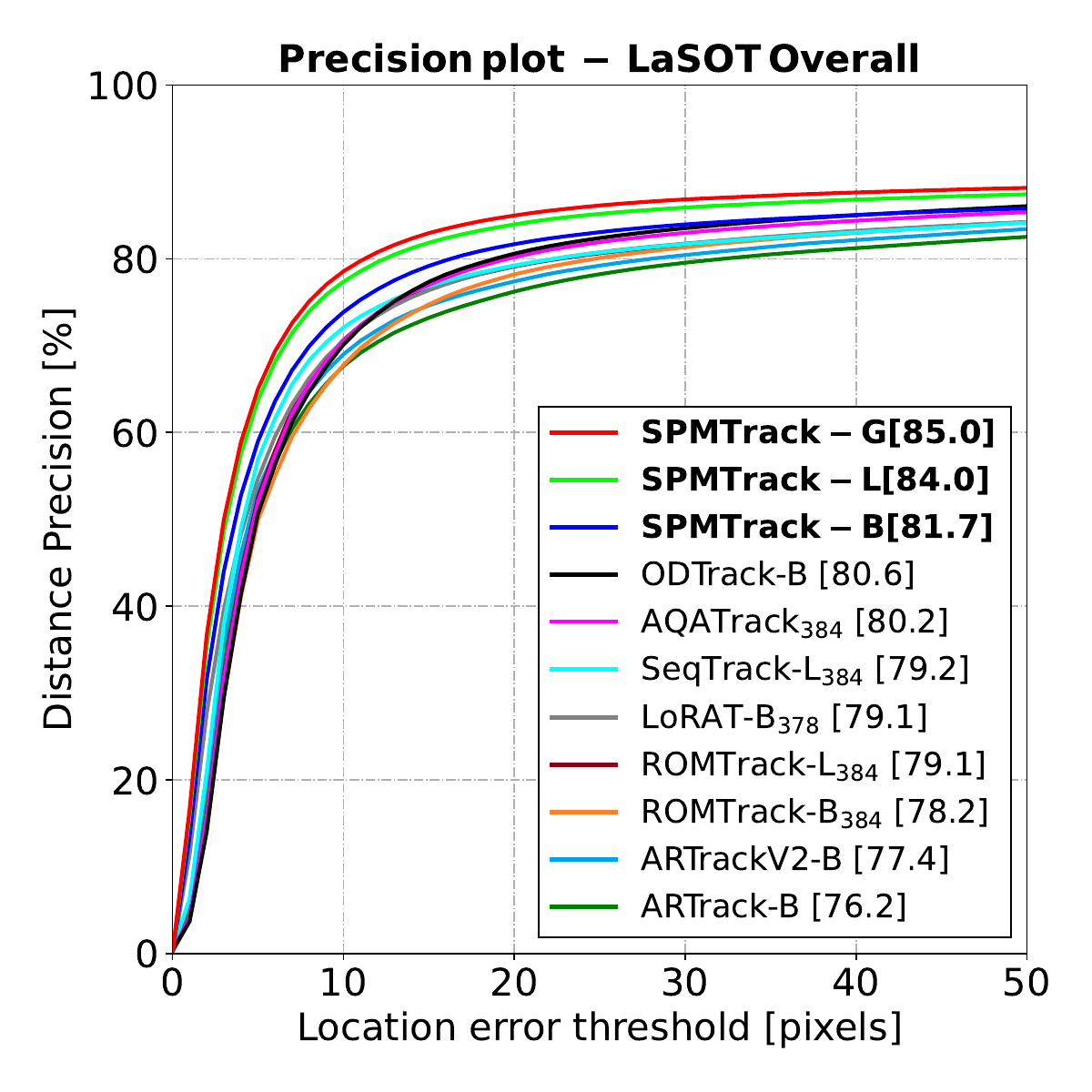}
\end{minipage}
}
}
\caption{Comparisons of our proposed SPMTrack with other excellent trackers in the success curve on LaSOT \emph{test} split, which includes eleven challenging scenarios such as Low Resolution, Motion Blur, Scale Variation, etc. We also provide the comparisons of the success and precision curves across the entire LaSOT \emph{test} split. Zoom in for better view.}
\label{fig:comparision of lasot dataset}
\vspace{-1ex}
\end{figure*}

\begin{figure*}[]
  \centering
    \vspace{3ex}
    \subfigure[Qualitative results of three methods when the targets undergo large deformations.]{\includegraphics[width=\textwidth]{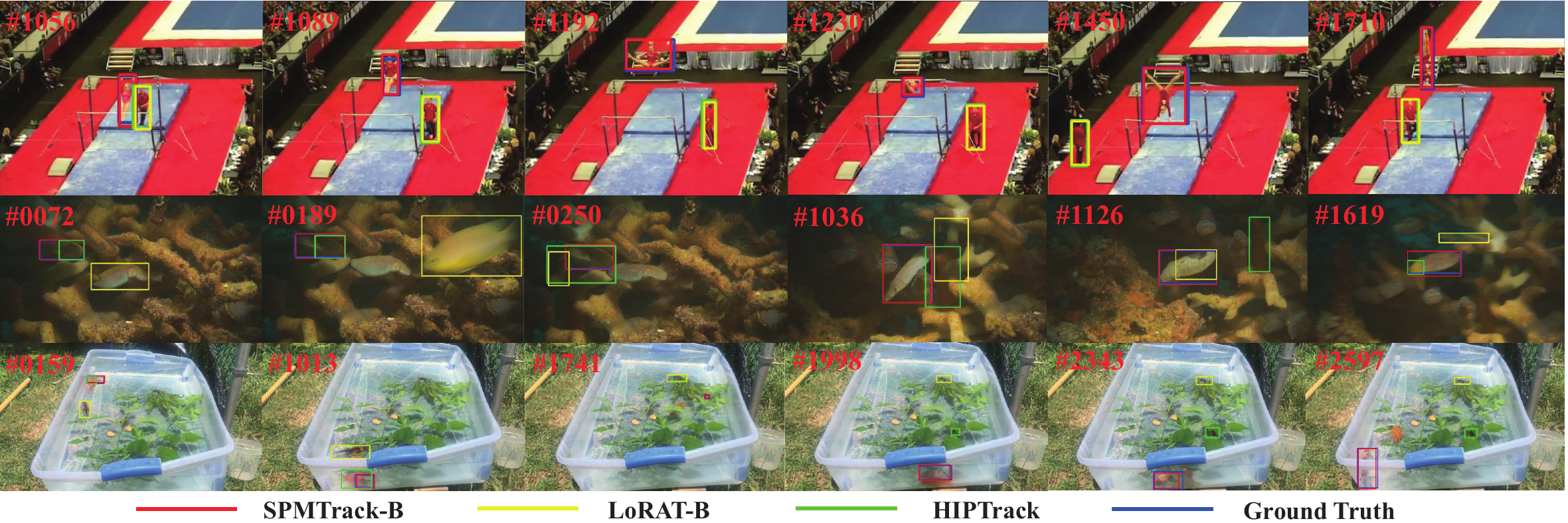}}\label{fig-vis-def} \\
    \subfigure[ Qualitative results of three methods when the targets suffer from partial occlusions.]
    {\includegraphics[width=\textwidth]{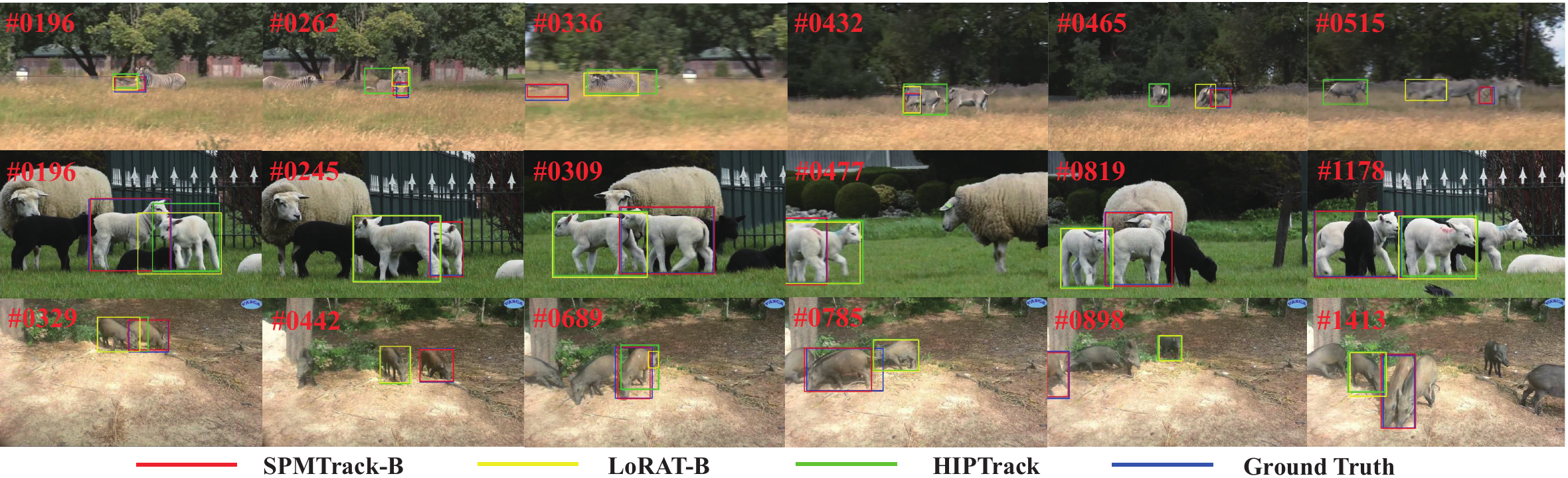}}\label{fig-vis-poc} \\
    \subfigure[ Qualitative results of three methods when the targets have large scale variations.]{\includegraphics[width=\textwidth]{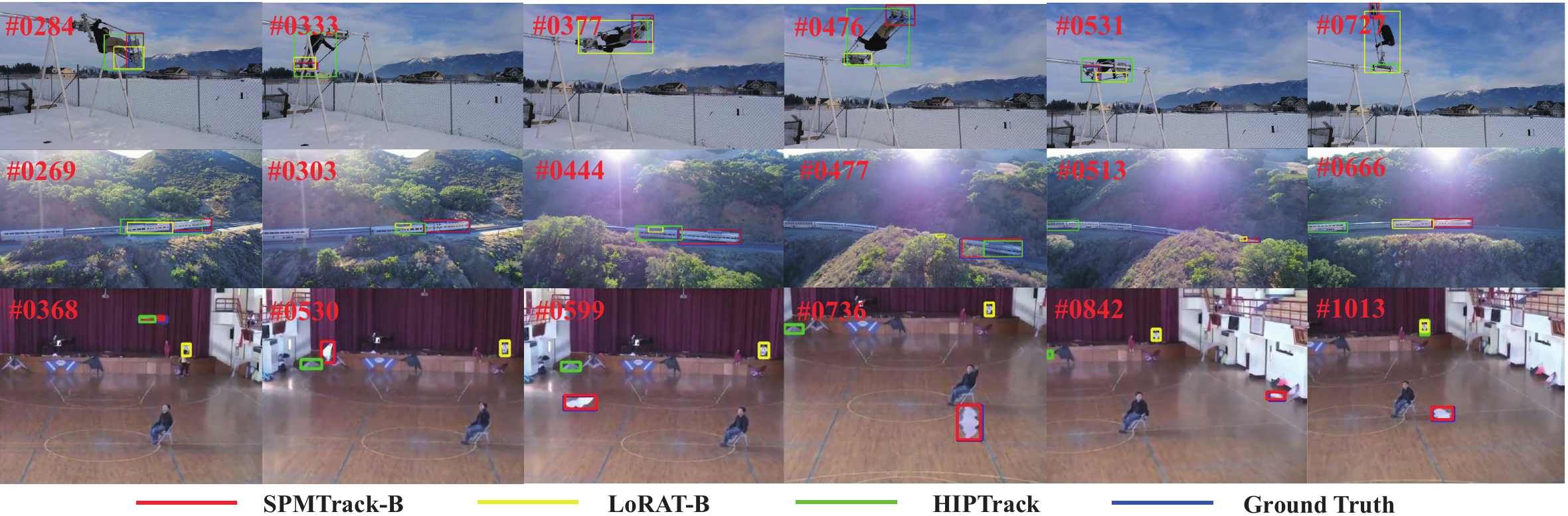}}\label{fig-vis-sv} \\
  \caption{ This figure presents a visual comparison among our proposed SPMTrack-B, LoRAT-B \cite{LoRAT} and HIPTrack \cite{Cai_2024_CVPR_HIPTrack} in the challenges of target deformation, partial occlusion and scale variation. It demonstrates that our method achieves more effective and accurate tracking in the aforementioned challenging scenarios. Zoom in for better view.}
    \label{fig:quantitative}
\end{figure*}

\begin{figure*}[!t]
\centering
\includegraphics[scale=0.5]{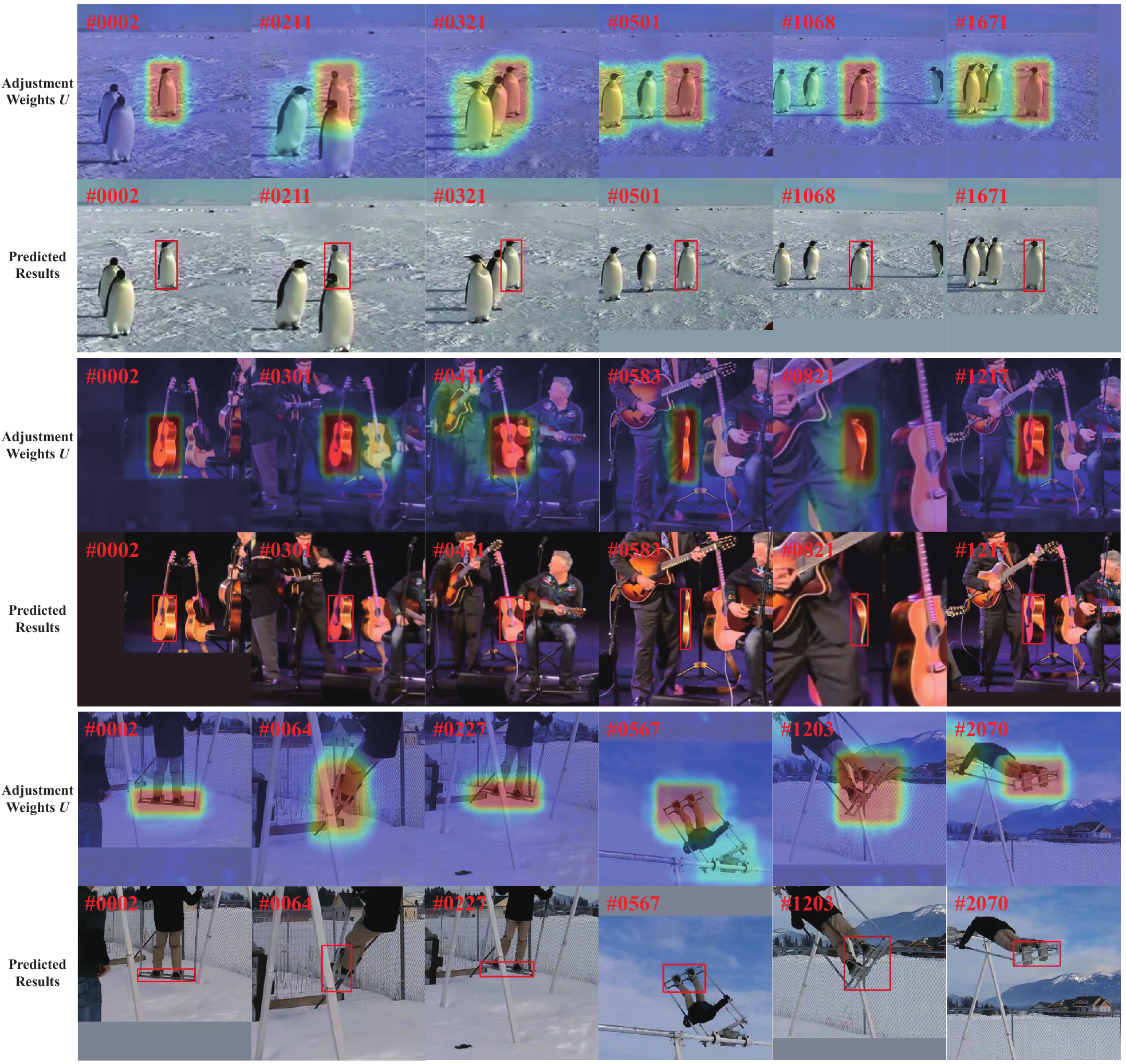}
\caption{ 
Visualization of the search region feature adjustment weights $\bm{U}$ and the corresponding predicted bounding boxes.
}
\label{fig:adjustments_weight}
\end{figure*}

\subsection{More Comparisons with Conventional MoE}

In Table \textcolor{cvprblue}{7} of the main manuscript, we compare our method with conventional MoE in terms of parameter count and performance on LaSOT, the conventional MoE structure in Table \textcolor{cvprblue}{7} is illustrated in Figure \ref{fig:conventional_moe}(a), most current large language models adopt a similar architecture~\cite{puigcerversparse_from_sparse_to_soft_moe,dai-etal-2024-deepseekmoe,wumixture_mixture_of_lora_experts,ma2018modeling_multigate_moe,fedus2022switchtransformers,clark2022unified_scaling_laws_for_routed}. Additionally, as shown in Figure \ref{fig:conventional_moe}, we explore various alternative MoE structures. 
Figure \ref{fig:conventional_moe}(b) shows a modification of the architecture in Figure \ref{fig:conventional_moe}(a) where we replace the frozen shared expert with a learnable routed expert and initialize all five routed experts with corresponding FFN weights from the pre-trained model. Figure \ref{fig:conventional_moe}(c) extends the design in Figure \ref{fig:conventional_moe}(a) by freezing all experts and fine-tuning each expert using LoRA \cite{hu_lora}. 
In Figure \ref{fig:conventional_moe}(d), we adopt the way of sparsely activating experts with top-k routing scores on the basis of Figure \ref{fig:conventional_moe}(a).
The experimental results of these different  MoE architectures on LaSOT \emph{test} split are presented in Table \ref{table_supp_conventional_moe}. 


As shown in Table \ref{table_supp_conventional_moe}, in row 1, the MoE comprises one shared expert and four routed experts, with all experts participating in computation, hence the expert count is $1+4$ as well as the activated expert count. Row 3 follows the same rule. In row 2, there are only five routed experts, and all of them participate in the computation. In  row 4, the MoE includes a total of one shared expert and four routed experts, but only the routed experts with the top-2 routing scores are activated, so the number of activated experts is $1 + 2$. In row 5, the MoE includes a total of one shared experts and eight routed experts, and the routed experts with the top-4 routing scores are activated.

Table \ref{table_supp_conventional_moe} demonstrates that TMoE achieves superior performance compared to all conventional MoE approaches that are only applied to replace FFN layers, while maintaining significantly lower total parameters (all conventional MoE variants exceed 300M parameters). At the same time, comparing row 2 with other rows, we observe that preserve pre-trained weights in MoE leads to substantially improved performance. This explains why row 3 achieves optimal performance among all conventional MoE approaches. Comparing rows 1, 4, and 5, we find that increasing the overall parameter count or the number of activated experts results in only marginal performance gains, further validating the effectiveness of our TMoE design.

The experimental results indicate that when using conventional MoE, the performance on visual tracking is inferior to that of TMoE. However, this does not mean that TMoE can achieve better performance on large language models. For example, the reinforcement learning~\cite{huang2026adaptivebatchwisesamplescheduling,huang2026realtimealignedrewardmodel,huang2026doesreasoningmodelimplicitly,zhang2026heterogeneousagentcollaborativereinforcement,ma2025stabilizingmoereinforcementlearning,zheng2025groupsequencepolicyoptimization,zheng2025stabilizingreinforcementlearningllms} phase of LLM requires routing replay to ensure stability, and our finer-grained expert design also leads to an increase in the number of routing operations, which raises the difficulty of training.

At the same time, in the field of natural language processing, sparsely activated MoE is the mainstream. But sparsely activated MoE requires more engineering optimizations. Due to the limitations of time and resources, we have not further explored the performance of TMoE under sparse activation. We also hope that this work can inspire other researchers to conduct related explorations in the field of visual tracking.

\begin{table}[h]\small
    \centering
    \caption{Performance comparison between different conventional MoE approaches and TMoE. The number of experts is represented in the form of $\bm{a} + \bm{b}$, where $\bm{a}$ represents shared experts and $\bm{b}$ represents routed experts. All results are evaluated on LaSOT \emph{test} split.}
    \setlength{\tabcolsep}{0.2mm}
    \begin{tabular}{c|cc|ccc}
    \Xhline{2pt}
        \textbf{MoE} & \textbf{\#MLP} & \textbf{\#Activated} & \multirow{2}{*}{AUC (\%)} & \multirow{2}{*}{\textbf{$P_{Norm} (\%)$}} & \multirow{2}{*}{\textbf{$P (\%)$}} \\
        \textbf{Variants} & \textbf{Experts} & \textbf{Experts} & & & \\
        \Xhline{0.5pt}
        Figure \ref{fig:conventional_moe}(a)  & 1+4 & 1+4 & 73.4 & 82.6 & 79.8 \\
        \Xhline{0.5pt}
        Figure \ref{fig:conventional_moe}(b)  & 5 & 5 & 72.7 & 81.5 & 78.7 \\
        \Xhline{0.5pt}
        Figure \ref{fig:conventional_moe}(c)  & 1+4 & 1+4 & 74.4 & 83.4 & 80.6 \\
        \Xhline{0.5pt}
        \multirow{2}{*}{Figure \ref{fig:conventional_moe}(d)}  & 1+4 & 1+2 & 73.3 & 82.0 & 79.4 \\
        & 1+8 & 1+4 & 73.8 & 82.8 & 79.9 \\
        \Xhline{0.5pt}
        \textbf{TMoE}  & - & - & \textbf{74.9} & \textbf{84.0} & \textbf{81.7} \\ 
    \Xhline{2pt}
    \end{tabular}
    \label{table_supp_conventional_moe}
\end{table}

\section{More Detailed Results in Different Attribute Scenes on LaSOT} \label{detailsinlasot}
In Figure \ref{fig:comparision of lasot dataset}, we provide a more detailed comparison of our method with other current state-of-the-art trackers LoRAT \cite{LoRAT}, ODTrack \cite{ODTrack}, ARTrackV2 \cite{Bai_2024_CVPR_ARTrackV2}, AQATrack \cite{Xie_2024_CVPR_AQATrack}, ARTrack \cite{Wei_2023_CVPR_autoregressive}, ROMTrack \cite{Cai_2023_ICCV_ROMTrack} and SeqTrack \cite{Chen_2023_CVPR_seqtrack} across various challenging scenario subsets in LaSOT \cite{fan2019lasot}.
Figure \ref{fig:comparision of lasot dataset} presents detailed success curves and AUC scores across individual subsets, along with the success and precision curves on the entire LaSOT \emph{test} split. The results demonstrate that SPMTrack-B significantly outperforms current state-of-the-art approaches both overall and across the vast majority of subsets.

\section{More Qualitative Results}\label{qualitative}

\subsection{Tracking Results}

In order to visually highlight the advantages of our method over existing approaches in challenging scenarios, we provide additional visualization results in Figure \ref{fig:quantitative}. All videos are from the \emph{test} split of LaSOT. We compare our proposed SPMTrack-B with HIPTrack \cite{Cai_2024_CVPR_HIPTrack} and LoRAT-B \cite{LoRAT} in terms of performance when the target undergoes deformation, occlusion, and scale variation. All the selected videos are challenging, as described below:

\begin{itemize}
    \item Figure \ref{fig:quantitative}(a) demonstrates the tracking results of three methods when the target suffers deformations.
    \item Figure \ref{fig:quantitative}(b) demonstrates the tracking results of three methods when the target suffers partial occlusions.
    \item Figure \ref{fig:quantitative}(c) demonstrates the tracking results of three methods when the target suffers scale variations.
\end{itemize}

\subsection{Visualization of Search Region Feature Adjustment Weight}
In the prediction head of SPMTrack, we compute matrix multiplication between the output $\bm{H}'$ corresponding to target state token and the output  search region feature $\bm{X}'$ to obtain weight $\bm{U}$. The obtained weight $\bm{U}$ contains the historical state information of the target and is used to further adjust and refine the output search region feature $\bm{X}'$. In Figure \ref{fig:adjustments_weight}, we visualize the weight $\bm{U}$. 
The visualization results demonstrate that the adjustment weight $\bm{U}$ can significantly distinguish between the target foreground region and irrelevant background regions. Moreover, the heatmaps exhibit patterns resembling bounding boxes, with substantially higher weights inside the ``bounding box". This further enhances the features at potential target locations, thereby improving the foreground-background discrimination capability of the search region features.


%
%
%

{
    \small
    \bibliographystyle{ieeenat_fullname}
    \bibliography{main}

@String(CVPR= {IEEE Conf. Comput. Vis. Pattern Recog.})

@String(ICCV= {Int. Conf. Comput. Vis.})

@String(ECCV= {Eur. Conf. Comput. Vis.})

@String(NIPS= {Adv. Neural Inform. Process. Syst.})

@String(ICLR = {Int. Conf. Learn. Represent.})

@String(AAAI = {AAAI})

@String(CVPR  = {CVPR})

@String(ICCV  = {ICCV})

@String(ECCV  = {ECCV})

@String(NIPS  = {NeurIPS})

@String(ICLR  = {ICLR})

@misc{ma2025stabilizingmoereinforcementlearning,
      title={Stabilizing MoE Reinforcement Learning by Aligning Training and Inference Routers}, 
      author={Wenhan Ma and Hailin Zhang and Liang Zhao and Yifan Song and Yudong Wang and Zhifang Sui and Fuli Luo},
      year={2025},
      eprint={2510.11370},
      archivePrefix={arXiv},
      primaryClass={cs.CL},
      url={https://arxiv.org/abs/2510.11370}, 
}

@misc{zheng2025groupsequencepolicyoptimization,
      title={Group Sequence Policy Optimization}, 
      author={Chujie Zheng and Shixuan Liu and Mingze Li and Xiong-Hui Chen and Bowen Yu and Chang Gao and Kai Dang and Yuqiong Liu and Rui Men and An Yang and Jingren Zhou and Junyang Lin},
      year={2025},
      eprint={2507.18071},
      archivePrefix={arXiv},
      primaryClass={cs.LG},
      url={https://arxiv.org/abs/2507.18071}, 
}

@misc{zheng2025stabilizingreinforcementlearningllms,
      title={Stabilizing Reinforcement Learning with LLMs: Formulation and Practices}, 
      author={Chujie Zheng and Kai Dang and Bowen Yu and Mingze Li and Huiqiang Jiang and Junrong Lin and Yuqiong Liu and Hao Lin and Chencan Wu and Feng Hu and An Yang and Jingren Zhou and Junyang Lin},
      year={2025},
      eprint={2512.01374},
      archivePrefix={arXiv},
      primaryClass={cs.LG},
      url={https://arxiv.org/abs/2512.01374}, 
}

@misc{huang2026doesreasoningmodelimplicitly,
      title={Does Your Reasoning Model Implicitly Know When to Stop Thinking?}, 
      author={Zixuan Huang and Xin Xia and Yuxi Ren and Jianbin Zheng and Xuanda Wang and Zhixia Zhang and Hongyan Xie and Songshi Liang and Zehao Chen and Xuefeng Xiao and Fuzhen Zhuang and Jianxin Li and Yikun Ban and Deqing Wang},
      year={2026},
      eprint={2602.08354},
      archivePrefix={arXiv},
      primaryClass={cs.AI},
      url={https://arxiv.org/abs/2602.08354}, 
}

@misc{huang2026realtimealignedrewardmodel,
      title={Real-Time Aligned Reward Model beyond Semantics}, 
      author={Zixuan Huang and Xin Xia and Yuxi Ren and Jianbin Zheng and Xuefeng Xiao and Hongyan Xie and Li Huaqiu and Songshi Liang and Zhongxiang Dai and Fuzhen Zhuang and Jianxin Li and Yikun Ban and Deqing Wang},
      year={2026},
      eprint={2601.22664},
      archivePrefix={arXiv},
      primaryClass={cs.AI},
      url={https://arxiv.org/abs/2601.22664}, 
}

@misc{huang2026adaptivebatchwisesamplescheduling,
      title={Adaptive Batch-Wise Sample Scheduling for Direct Preference Optimization}, 
      author={Zixuan Huang and Yikun Ban and Lean Fu and Xiaojie Li and Zhongxiang Dai and Jianxin Li and Deqing Wang},
      year={2026},
      eprint={2506.17252},
      archivePrefix={arXiv},
      primaryClass={cs.LG},
      url={https://arxiv.org/abs/2506.17252}, 
}

@misc{zhang2026heterogeneousagentcollaborativereinforcement,
      title={Heterogeneous Agent Collaborative Reinforcement Learning}, 
      author={Zhixia Zhang and Zixuan Huang and Xin Xia and Deqing Wang and Fuzhen Zhuang and Shuai Ma and Ning Ding and Yaodong Yang and Jianxin Li and Yikun Ban},
      year={2026},
      eprint={2603.02604},
      archivePrefix={arXiv},
      primaryClass={cs.LG},
      url={https://arxiv.org/abs/2603.02604}, 
}

@article{wang2024survey_data_synthesis_ours,
  title={A survey on data synthesis and augmentation for large language models},
  author={Wang, Ke and Zhu, Jiahui and Ren, Minjie and Liu, Zeming and Li, Shiwei and Zhang, Zongye and Zhang, Chenkai and Wu, Xiaoyu and Zhan, Qiqi and Liu, Qingjie and others},
  journal={arXiv preprint arXiv:2410.12896},
  year={2024}
}

@inproceedings{ma2018modeling_multigate_moe,
  title={Modeling task relationships in multi-task learning with multi-gate mixture-of-experts},
  author={Ma, Jiaqi and Zhao, Zhe and Yi, Xinyang and Chen, Jilin and Hong, Lichan and Chi, Ed H},
  booktitle={Proceedings of the 24th ACM SIGKDD international conference on knowledge discovery \& data mining},
  pages={1930--1939},
  year={2018}
}

@inproceedings{houlsby2019parameter_adapter,
  title={Parameter-efficient transfer learning for NLP},
  author={Houlsby, Neil and Giurgiu, Andrei and Jastrzebski, Stanislaw and Morrone, Bruna and De Laroussilhe, Quentin and Gesmundo, Andrea and Attariyan, Mona and Gelly, Sylvain},
  booktitle={International conference on machine learning},
  pages={2790--2799},
  year={2019},
  organization={PMLR}
}

@article{oquab2024dinov2,
  title={DINOv2: Learning Robust Visual Features without Supervision},
  author={Oquab, Maxime and Darcet, Timoth{\'e}e and Moutakanni, Th{\'e}o and Vo, Huy and Szafraniec, Marc and Khalidov, Vasil and Fernandez, Pierre and Haziza, Daniel and Massa, Francisco and El-Nouby, Alaaeldin and others},
  journal={Transactions on Machine Learning Research Journal},
  pages={1--31},
  year={2024}
}

@inproceedings{DBLP:conf/iclr/LoshchilovH19_AdamW,
  author       = {Ilya Loshchilov and
                  Frank Hutter},
  title        = {Decoupled Weight Decay Regularization},
  booktitle    = {7th International Conference on Learning Representations, {ICLR}},
  year         = {2019},
  bibsource    = {dblp computer science bibliography, https://dblp.org}
}

@inproceedings{pfeiffer2020AdapterHub_mad_x,
    title={AdapterHub: A Framework for Adapting Transformers},
    author={Pfeiffer, Jonas and
            R{\"u}ckl{\'e}, Andreas and
            Poth, Clifton and
            Kamath, Aishwarya and
            Vuli{\'c}, Ivan and
            Ruder, Sebastian and
            Cho, Kyunghyun and
            Gurevych, Iryna},
    booktitle={Proceedings of the 2020 Conference on Empirical Methods in Natural Language Processing: System Demonstrations},
    pages={46--54},
    year={2020}
}

@inproceedings{tian2024hydralora,
  title={HydraLoRA: An Asymmetric LoRA Architecture for Efficient Fine-Tuning},
  author={Tian, Chunlin and Shi, Zhan and Guo, Zhijiang and Li, Li and Xu, Chengzhong},
  booktitle={Advances in Neural Information Processing Systems (NeurIPS)},
  year={2024}
}

@inproceedings{hu_lora,
  author       = {Edward J. Hu and
                  Yelong Shen and
                  Phillip Wallis and
                  Zeyuan Allen{-}Zhu and
                  Yuanzhi Li and
                  Shean Wang and
                  Lu Wang and
                  Weizhu Chen},
  title        = {LoRA: Low-Rank Adaptation of Large Language Models},
  booktitle    = {The Tenth International Conference on Learning Representations, {ICLR}
                },
  year         = {2022},
  url          = {https://openreview.net/forum?id=nZeVKeeFYf9},
  bibsource    = {dblp computer science bibliography, https://dblp.org}
}

@article{gao2024clip-adapter,
  title={Clip-adapter: Better vision-language models with feature adapters},
  author={Gao, Peng and Geng, Shijie and Zhang, Renrui and Ma, Teli and Fang, Rongyao and Zhang, Yongfeng and Li, Hongsheng and Qiao, Yu},
  journal={International Journal of Computer Vision},
  volume={132},
  number={2},
  pages={581--595},
  year={2024},
  publisher={Springer}
}

@article{chen2024emoetracker,
  title={eMoE-Tracker: Environmental MoE-based Transformer for Robust Event-guided Object Tracking},
  author={Chen, Yucheng and Wang, Lin},
  journal={arXiv preprint arXiv:2406.20024},
  year={2024}
}

@article{tang2024revisiting_moetrack,
  title={Revisiting RGBT Tracking Benchmarks from the Perspective of Modality Validity: A New Benchmark, Problem, and Method},
  author={Tang, Zhangyong and Xu, Tianyang and Feng, Zhenhua and Zhu, Xuefeng and Wang, He and Shao, Pengcheng and Cheng, Chunyang and Wu, Xiao-Jun and Awais, Muhammad and Atito, Sara and others},
  journal={arXiv preprint arXiv:2405.00168},
  year={2024}
}

@InProceedings{pmlr-v202-chowdhury23a-patch-level-routing-in-moe,
  title = 	 {Patch-level Routing in Mixture-of-Experts is Provably Sample-efficient for Convolutional Neural Networks},
  author =       {Chowdhury, Mohammed Nowaz Rabbani and Zhang, Shuai and Wang, Meng and Liu, Sijia and Chen, Pin-Yu},
  booktitle = 	 {Proceedings of the 40th International Conference on Machine Learning},
  pages = 	 {6074--6114},
  year = 	 {2023},
  editor = 	 {Krause, Andreas and Brunskill, Emma and Cho, Kyunghyun and Engelhardt, Barbara and Sabato, Sivan and Scarlett, Jonathan},
  volume = 	 {202},
  series = 	 {Proceedings of Machine Learning Research},
  month = 	 {23--29 Jul},
  publisher =    {PMLR},
  url = 	 {https://proceedings.mlr.press/v202/chowdhury23a.html}
}

@InProceedings{He_2022_CVPR_mae,
    author    = {He, Kaiming and Chen, Xinlei and Xie, Saining and Li, Yanghao and Doll\'ar, Piotr and Girshick, Ross},
    title     = {Masked Autoencoders Are Scalable Vision Learners},
    booktitle = {Proceedings of the IEEE/CVF Conference on Computer Vision and Pattern Recognition (CVPR)},
    month     = {June},
    year      = {2022},
    pages     = {16000-16009}
}

@InProceedings{Zhang_2023_ICCV_Robust_mox_convolutional,
    author    = {Zhang, Yihua and Cai, Ruisi and Chen, Tianlong and Zhang, Guanhua and Zhang, Huan and Chen, Pin-Yu and Chang, Shiyu and Wang, Zhangyang and Liu, Sijia},
    title     = {Robust Mixture-of-Expert Training for Convolutional Neural Networks},
    booktitle = {Proceedings of the IEEE/CVF International Conference on Computer Vision (ICCV)},
    month     = {October},
    year      = {2023},
    pages     = {90-101}
}

@inproceedings{NEURIPS2021_48237d9f_scaling_vision_sparse_moe,
 author = {Riquelme, Carlos and Puigcerver, Joan and Mustafa, Basil and Neumann, Maxim and Jenatton, Rodolphe and Susano Pinto, Andr\'{e} and Keysers, Daniel and Houlsby, Neil},
 booktitle = {Advances in Neural Information Processing Systems},
 editor = {M. Ranzato and A. Beygelzimer and Y. Dauphin and P.S. Liang and J. Wortman Vaughan},
 pages = {8583--8595},
 publisher = {Curran Associates, Inc.},
 title = {Scaling Vision with Sparse Mixture of Experts},
 url = {https://proceedings.neurips.cc/paper_files/paper/2021/file/48237d9f2dea8c74c2a72126cf63d933-Paper.pdf},
 volume = {34},
 year = {2021}
}

@inproceedings{clark2022unified_scaling_laws_for_routed,
  title={Unified scaling laws for routed language models},
  author={Clark, Aidan and de Las Casas, Diego and Guy, Aurelia and Mensch, Arthur and Paganini, Michela and Hoffmann, Jordan and Damoc, Bogdan and Hechtman, Blake and Cai, Trevor and Borgeaud, Sebastian and others},
  booktitle={International conference on machine learning},
  pages={4057--4086},
  year={2022},
  organization={PMLR}
}

@article{fedus2022switchtransformers,
  title={Switch transformers: Scaling to trillion parameter models with simple and efficient sparsity},
  author={Fedus, William and Zoph, Barret and Shazeer, Noam},
  journal={Journal of Machine Learning Research},
  volume={23},
  number={120},
  pages={1--39},
  year={2022}
}

@inproceedings{wumixture_mixture_of_lora_experts,
  author       = {Xun Wu and
                  Shaohan Huang and
                  Furu Wei},
  title        = {Mixture of LoRA Experts},
  booktitle    = {The Twelfth International Conference on Learning Representations,
                  {ICLR}},
  year         = {2024},
  url          = {https://openreview.net/forum?id=uWvKBCYh4S},
  bibsource    = {dblp computer science bibliography, https://dblp.org}
}

@inproceedings{dai-etal-2024-deepseekmoe,
    title = "{D}eep{S}eek{M}o{E}: Towards Ultimate Expert Specialization in Mixture-of-Experts Language Models",
    author = "Dai, Damai  and
      Deng, Chengqi  and
      Zhao, Chenggang  and
      Xu, R.x.  and
      Gao, Huazuo  and
      Chen, Deli  and
      Li, Jiashi  and
      Zeng, Wangding  and
      Yu, Xingkai  and
      Wu, Y.  and
      Xie, Zhenda  and
      Li, Y.k.  and
      Huang, Panpan  and
      Luo, Fuli  and
      Ruan, Chong  and
      Sui, Zhifang  and
      Liang, Wenfeng",
    editor = "Ku, Lun-Wei  and
      Martins, Andre  and
      Srikumar, Vivek",
    booktitle = "Proceedings of the 62nd Annual Meeting of the Association for Computational Linguistics (Volume 1: Long Papers)",
    month = aug,
    year = "2024",
    address = "Bangkok, Thailand",
    publisher = "Association for Computational Linguistics",
    url = "https://aclanthology.org/2024.acl-long.70",
    doi = "10.18653/v1/2024.acl-long.70",
    pages = "1280--1297",
}

@inproceedings{puigcerversparse_from_sparse_to_soft_moe,
  author       = {Joan Puigcerver and
                  Carlos Riquelme Ruiz and
                  Basil Mustafa and
                  Neil Houlsby},
  title        = {From Sparse to Soft Mixtures of Experts},
  booktitle    = {The Twelfth International Conference on Learning Representations,
                  {ICLR}},
  year         = {2024},
  url          = {https://openreview.net/forum?id=jxpsAj7ltE},
  bibsource    = {dblp computer science bibliography, https://dblp.org}
}

@InProceedings{Aljundi_2017_CVPR_expert_gate,
author = {Aljundi, Rahaf and Chakravarty, Punarjay and Tuytelaars, Tinne},
title = {Expert Gate: Lifelong Learning With a Network of Experts},
booktitle = {Proceedings of the IEEE Conference on Computer Vision and Pattern Recognition (CVPR)},
month = {July},
year = {2017}
}

@InProceedings{Cai_2023_ICCV_ROMTrack,
    author    = {Cai, Yidong and Liu, Jie and Tang, Jie and Wu, Gangshan},
    title     = {Robust Object Modeling for Visual Tracking},
    booktitle = {Proceedings of the IEEE/CVF International Conference on Computer Vision (ICCV)},
    month     = {October},
    year      = {2023},
    pages     = {9589-9600}
}

@inproceedings{NEURIPS2022_1c8c87c3_VLT,
 author = {Guo, Mingzhe and Zhang, Zhipeng and Fan, Heng and Jing, Liping},
 booktitle = {Advances in Neural Information Processing Systems},
 editor = {S. Koyejo and S. Mohamed and A. Agarwal and D. Belgrave and K. Cho and A. Oh},
 pages = {4446--4460},
 publisher = {Curran Associates, Inc.},
 title = {Divert More Attention to Vision-Language Tracking},
 url = {https://proceedings.neurips.cc/paper_files/paper/2022/file/1c8c87c36dc1e49e63555f95fa56b153-Paper-Conference.pdf},
 volume = {35},
 year = {2022}
}

@InProceedings{Yan_2023_CVPR_UNINEXT,
    author    = {Yan, Bin and Jiang, Yi and Wu, Jiannan and Wang, Dong and Luo, Ping and Yuan, Zehuan and Lu, Huchuan},
    title     = {Universal Instance Perception As Object Discovery and Retrieval},
    booktitle = {Proceedings of the IEEE/CVF Conference on Computer Vision and Pattern Recognition (CVPR)},
    month     = {June},
    year      = {2023},
    pages     = {15325-15336}
}

@InProceedings{Li_2023_ICCV_citetracker,
    author    = {Li, Xin and Huang, Yuqing and He, Zhenyu and Wang, Yaowei and Lu, Huchuan and Yang, Ming-Hsuan},
    title     = {CiteTracker: Correlating Image and Text for Visual Tracking},
    booktitle = {Proceedings of the IEEE/CVF International Conference on Computer Vision (ICCV)},
    month     = {October},
    year      = {2023},
    pages     = {9974-9983}
}

@InProceedings{Huang_2024_CVPR_RTracker,
    author    = {Huang, Yuqing and Li, Xin and Zhou, Zikun and Wang, Yaowei and He, Zhenyu and Yang, Ming-Hsuan},
    title     = {RTracker: Recoverable Tracking via PN Tree Structured Memory},
    booktitle = {Proceedings of the IEEE/CVF Conference on Computer Vision and Pattern Recognition (CVPR)},
    month     = {June},
    year      = {2024},
    pages     = {19038-19047}
}

@InProceedings{Xie_2024_CVPR_AQATrack,
    author    = {Xie, Jinxia and Zhong, Bineng and Mo, Zhiyi and Zhang, Shengping and Shi, Liangtao and Song, Shuxiang and Ji, Rongrong},
    title     = {Autoregressive Queries for Adaptive Tracking with Spatio-Temporal Transformers},
    booktitle = {Proceedings of the IEEE/CVF Conference on Computer Vision and Pattern Recognition (CVPR)},
    month     = {June},
    year      = {2024},
    pages     = {19300-19309}
}

@InProceedings{Bai_2024_CVPR_ARTrackV2,
    author    = {Bai, Yifan and Zhao, Zeyang and Gong, Yihong and Wei, Xing},
    title     = {ARTrackV2: Prompting Autoregressive Tracker Where to Look and How to Describe},
    booktitle = {Proceedings of the IEEE/CVF Conference on Computer Vision and Pattern Recognition (CVPR)},
    month     = {June},
    year      = {2024},
    pages     = {19048-19057}
}

@InProceedings{Wang_2021_CVPR_TNL2k,
    author    = {Wang, Xiao and Shu, Xiujun and Zhang, Zhipeng and Jiang, Bo and Wang, Yaowei and Tian, Yonghong and Wu, Feng},
    title     = {Towards More Flexible and Accurate Object Tracking With Natural Language: Algorithms and Benchmark},
    booktitle = {Proceedings of the IEEE/CVF Conference on Computer Vision and Pattern Recognition (CVPR)},
    month     = {June},
    year      = {2021},
    pages     = {13763-13773}
}

@InProceedings{pmlr-v162-du22c_moe2,
  title = 	 {{GL}a{M}: Efficient Scaling of Language Models with Mixture-of-Experts},
  author =       {Du, Nan and Huang, Yanping and Dai, Andrew M and Tong, Simon and Lepikhin, Dmitry and Xu, Yuanzhong and Krikun, Maxim and Zhou, Yanqi and Yu, Adams Wei and Firat, Orhan and Zoph, Barret and Fedus, Liam and Bosma, Maarten P and Zhou, Zongwei and Wang, Tao and Wang, Emma and Webster, Kellie and Pellat, Marie and Robinson, Kevin and Meier-Hellstern, Kathleen and Duke, Toju and Dixon, Lucas and Zhang, Kun and Le, Quoc and Wu, Yonghui and Chen, Zhifeng and Cui, Claire},
  booktitle = 	 {Proceedings of the 39th International Conference on Machine Learning},
  pages = 	 {5547--5569},
  year = 	 {2022},
  editor = 	 {Chaudhuri, Kamalika and Jegelka, Stefanie and Song, Le and Szepesvari, Csaba and Niu, Gang and Sabato, Sivan},
  volume = 	 {162},
  series = 	 {Proceedings of Machine Learning Research},
  month = 	 {17--23 Jul},
  publisher =    {PMLR},
  url = 	 {https://proceedings.mlr.press/v162/du22c.html}
}

@article{jacobs1991adaptive_moe,
  title={Adaptive mixtures of local experts},
  author={Jacobs, Robert A and Jordan, Michael I and Nowlan, Steven J and Hinton, Geoffrey E},
  journal={Neural computation},
  volume={3},
  number={1},
  pages={79--87},
  year={1991},
  publisher={MIT Press}
}

@InProceedings{Yang_2023_ICCV_FBDMTrack,
    author    = {Yang, Dawei and He, Jianfeng and Ma, Yinchao and Yu, Qianjin and Zhang, Tianzhu},
    title     = {Foreground-Background Distribution Modeling Transformer for Visual Object Tracking},
    booktitle = {Proceedings of the IEEE/CVF International Conference on Computer Vision (ICCV)},
    month     = {October},
    year      = {2023},
    pages     = {10117-10127}
}

@InProceedings{Wei_2023_CVPR_autoregressive,
    author    = {Wei, Xing and Bai, Yifan and Zheng, Yongchao and Shi, Dahu and Gong, Yihong},
    title     = {Autoregressive Visual Tracking},
    booktitle = {Proceedings of the IEEE/CVF Conference on Computer Vision and Pattern Recognition (CVPR)},
    month     = {June},
    year      = {2023},
    pages     = {9697-9706}
}

@InProceedings{Chen_2023_CVPR_seqtrack,
    author    = {Chen, Xin and Peng, Houwen and Wang, Dong and Lu, Huchuan and Hu, Han},
    title     = {SeqTrack: Sequence to Sequence Learning for Visual Object Tracking},
    booktitle = {Proceedings of the IEEE/CVF Conference on Computer Vision and Pattern Recognition (CVPR)},
    month     = {June},
    year      = {2023},
    pages     = {14572-14581}
}

@InProceedings{Cui_2022_MixFormer,
    author    = {Cui, Yutao and Jiang, Cheng and Wang, Limin and Wu, Gangshan},
    title     = {MixFormer: End-to-End Tracking With Iterative Mixed Attention},
    booktitle = {Proceedings of the IEEE/CVF Conference on Computer Vision and Pattern Recognition (CVPR)},
    month     = {June},
    year      = {2022},
    pages     = {13608-13618}
}

@article{ODTrack, title={ODTrack: Online Dense Temporal Token Learning for Visual Tracking}, volume={38}, url={https://ojs.aaai.org/index.php/AAAI/article/view/28591}, DOI={10.1609/aaai.v38i7.28591}, abstractNote={Online contextual reasoning and association across consecutive video frames are critical to perceive instances in visual tracking. However, most current top-performing trackers persistently lean on sparse temporal relationships between reference and search frames via an offline mode. Consequently, they can only interact independently within each image-pair and establish limited temporal correlations. To alleviate the above problem, we propose a simple, flexible and effective video-level tracking pipeline, named ODTrack, which densely associates the contextual relationships of video frames in an online token propagation manner. ODTrack receives video frames of arbitrary length to capture the spatio-temporal trajectory relationships of an instance, and compresses the discrimination features (localization information) of a target into a token sequence to achieve frame-to-frame association. This new solution brings the following benefits: 1) the purified token sequences can serve as prompts for the inference in the next video frame, whereby past information is leveraged to guide future inference; 2) the complex online update strategies are effectively avoided by the iterative propagation of token sequences, and thus we can achieve more efficient model representation and computation. ODTrack achieves a new SOTA performance on seven benchmarks, while running at real-time speed. Code and models are available at https://github.com/GXNU-ZhongLab/ODTrack.}, number={7}, journal={Proceedings of the AAAI Conference on Artificial Intelligence}, author={Zheng, Yaozong and Zhong, Bineng and Liang, Qihua and Mo, Zhiyi and Zhang, Shengping and Li, Xianxian}, year={2024}, month={Mar.}, pages={7588-7596} }

@InProceedings{LoRAT,
author="Lin, Liting
and Fan, Heng
and Zhang, Zhipeng
and Wang, Yaowei
and Xu, Yong
and Ling, Haibin",
editor="Leonardis, Ale{\v{s}}
and Ricci, Elisa
and Roth, Stefan
and Russakovsky, Olga
and Sattler, Torsten
and Varol, G{\"u}l",
title="Tracking Meets LoRA: Faster Training, Larger Model, Stronger Performance",
booktitle="Computer Vision -- ECCV 2024",
year="2025",
publisher="Springer Nature Switzerland",
address="Cham",
pages="300--318",
isbn="978-3-031-73232-4"
}

@InProceedings{Cai_2024_CVPR_HIPTrack,
    author    = {Cai, Wenrui and Liu, Qingjie and Wang, Yunhong},
    title     = {HIPTrack: Visual Tracking with Historical Prompts},
    booktitle = {Proceedings of the IEEE/CVF Conference on Computer Vision and Pattern Recognition (CVPR)},
    month     = {June},
    year      = {2024},
    pages     = {19258-19267}
}

@inproceedings{fu2022sparsett,
  title={SparseTT: Visual tracking with sparse transformers},
  author={Fu, Zhihong and Fu, Zehua and Liu, Qingjie and Cai, Wenrui and Wang, Yunhong},
  month={July},
  year={2022},
  booktitle = {Proceedings of the Thirtieth International Joint Conference on Artificial Intelligence, {IJCAI-22}},
  pages={905-912}
}

@article{lin2022swintrack,
  title={Swintrack: A simple and strong baseline for transformer tracking},
  author={Lin, Liting and Fan, Heng and Zhang, Zhipeng and Xu, Yong and Ling, Haibin},
  journal={Advances in Neural Information Processing Systems},
  volume={35},
  pages={16743--16754},
  year={2022}
}

@inproceedings{chen2022backbone,
  title={Backbone is all your need: a simplified architecture for visual object tracking},
  author={Chen, Boyu and Li, Peixia and Bai, Lei and Qiao, Lei and Shen, Qiuhong and Li, Bo and Gan, Weihao and Wu, Wei and Ouyang, Wanli},
  booktitle={Computer Vision--ECCV 2022: 17th European Conference, Tel Aviv, Israel, October 23--27, 2022, Proceedings, Part XXII},
  pages={375--392},
  year={2022},
  organization={Springer}
}

@inproceedings{gao2022aiatrack,
  title={Aiatrack: Attention in attention for transformer visual tracking},
  author={Gao, Shenyuan and Zhou, Chunluan and Ma, Chao and Wang, Xinggang and Yuan, Junsong},
  booktitle={Computer Vision--ECCV 2022: 17th European Conference, Tel Aviv, Israel, October 23--27, 2022, Proceedings, Part XXII},
  pages={146--164},
  year={2022},
  organization={Springer}
}

@inproceedings{ye_2022_joint,
  title={Joint feature learning and relation modeling for tracking: A one-stream framework},
  author={Ye, Botao and Chang, Hong and Ma, Bingpeng and Shan, Shiguang and Chen, Xilin},
  booktitle={European Conference on Computer Vision},
  pages={341--357},
  year={2022},
  organization={Springer}
}

@InProceedings{Xie_2022_Correlation,
    author    = {Xie, Fei and Wang, Chunyu and Wang, Guangting and Cao, Yue and Yang, Wankou and Zeng, Wenjun},
    title     = {Correlation-Aware Deep Tracking},
    booktitle = {Proceedings of the IEEE/CVF Conference on Computer Vision and Pattern Recognition (CVPR)},
    month     = {June},
    year      = {2022},
    pages     = {8751-8760}
}

@inproceedings{wang2020tracking,
  title={Tracking by instance detection: A meta-learning approach},
  author={Wang, Guangting and Luo, Chong and Sun, Xiaoyan and Xiong, Zhiwei and Zeng, Wenjun},
  booktitle={CVPR},
  pages={6288--6297},
  year={2020}
}

@inproceedings{Vaswani2017AttentionIA,
	title={Attention is all you need},
	author={Vaswani, Ashish and Shazeer, Noam and Parmar, Niki and Uszkoreit, Jakob and Jones, Llion and Gomez, Aidan N and Kaiser, {\L}ukasz and Polosukhin, Illia},
	booktitle={NIPS},
	pages={5998--6008},
	year={2017}
}

@inproceedings{wang2022learningtopromptforcontinual,
  title={Learning to prompt for continual learning},
  author={Wang, Zifeng and Zhang, Zizhao and Lee, Chen-Yu and Zhang, Han and Sun, Ruoxi and Ren, Xiaoqi and Su, Guolong and Perot, Vincent and Dy, Jennifer and Pfister, Tomas},
  booktitle={Proceedings of the IEEE/CVF Conference on Computer Vision and Pattern Recognition},
  pages={139--149},
  year={2022}
}

@inproceedings{jia2022visualPromptTuning,
  title={Visual prompt tuning},
  author={Jia, Menglin and Tang, Luming and Chen, Bor-Chun and Cardie, Claire and Belongie, Serge and Hariharan, Bharath and Lim, Ser-Nam},
  booktitle={European Conference on Computer Vision},
  pages={709--727},
  year={2022},
  organization={Springer}
}

@article{zhou2022learning_to_prompt,
  title={Learning to prompt for vision-language models},
  author={Zhou, Kaiyang and Yang, Jingkang and Loy, Chen Change and Liu, Ziwei},
  journal={International Journal of Computer Vision},
  volume={130},
  number={9},
  pages={2337--2348},
  year={2022},
  publisher={Springer}
}

@inproceedings{li2021prefix,
  title={Prefix-Tuning: Optimizing Continuous Prompts for Generation},
  author={Li, Xiang Lisa and Liang, Percy},
  booktitle={Proceedings of the 59th Annual Meeting of the Association for Computational Linguistics and the 11th International Joint Conference on Natural Language Processing (Volume 1: Long Papers)},
  pages={4582--4597},
  year={2021}
}

@inproceedings{fu2021stmtrack,
  title={STMTrack: Template-free Visual Tracking with Space-time Memory Networks},
  author={Fu, Zhihong and Liu, Qingjie and Fu, Zehua and Wang, Yunhong},
  booktitle={CVPR},
  pages={13774--13783},
  year={2021}
}

@article{he2023target_TATrack, title={Target-Aware Tracking with Long-Term Context Attention}, volume={37}, url={https://ojs.aaai.org/index.php/AAAI/article/view/25155}, DOI={10.1609/aaai.v37i1.25155}, abstractNote={Most deep trackers still follow the guidance of the siamese paradigms and use a template that contains only the target without any contextual information, which makes it difficult for the tracker to cope with large appearance changes, rapid target movement, and attraction from similar objects. To alleviate the above problem, we propose a long-term context attention (LCA) module that can perform extensive information fusion on the target and its context from long-term frames, and calculate the target correlation while enhancing target features. The complete contextual information contains the location of the target as well as the state around the target. LCA uses the target state from the previous frame to exclude the interference of similar objects and complex backgrounds, thus accurately locating the target and enabling the tracker to obtain higher robustness and regression accuracy. By embedding the LCA module in Transformer, we build a powerful online tracker with a target-aware backbone, termed as TATrack. In addition, we propose a dynamic online update algorithm based on the classification confidence of historical information without additional calculation burden. Our tracker achieves state-of-the-art performance on multiple benchmarks, with 71.1% AUC, 89.3% NP, and 73.0% AO on LaSOT, TrackingNet, and GOT-10k. The code and trained models are available on https://github.com/hekaijie123/TATrack.}, number={1}, journal={Proceedings of the AAAI Conference on Artificial Intelligence}, author={He, Kaijie and Zhang, Canlong and Xie, Sheng and Li, Zhixin and Wang, Zhiwen}, year={2023}, month={Jun.}, pages={773-780} }

@InProceedings{Gao_2023_CVPR_GRM,
    author    = {Gao, Shenyuan and Zhou, Chunluan and Zhang, Jun},
    title     = {Generalized Relation Modeling for Transformer Tracking},
    booktitle = {Proceedings of the IEEE/CVF Conference on Computer Vision and Pattern Recognition (CVPR)},
    month     = {June},
    year      = {2023},
    pages     = {18686-18695}
}

@InProceedings{Galoogahi_2017_ICCV_nfs,
author = {Kiani Galoogahi, Hamed and Fagg, Ashton and Huang, Chen and Ramanan, Deva and Lucey, Simon},
title = {Need for Speed: A Benchmark for Higher Frame Rate Object Tracking},
booktitle = {Proceedings of the IEEE International Conference on Computer Vision (ICCV)},
month = {Oct},
year = {2017}
}

@inproceedings{rezatofighi2019generalized,
  title={Generalized intersection over union: A metric and a loss for bounding box regression},
  author={Rezatofighi, Hamid and Tsoi, Nathan and Gwak, JunYoung and Sadeghian, Amir and Reid, Ian and Savarese, Silvio},
  booktitle={CVPR},
  pages={658--666},
  year={2019}
}

@inproceedings{dosovitskiy2020image,
  title={An image is worth 16x16 words: Transformers for image recognition at scale},
  author={Dosovitskiy, Alexey and Beyer, Lucas and Kolesnikov, Alexander and Weissenborn, Dirk and Zhai, Xiaohua and Unterthiner, Thomas and Dehghani, Mostafa and Minderer, Matthias and Heigold, Georg and Gelly, Sylvain and others},
  booktitle={ICLR},
  year={2021}
}

@inproceedings{muller2018trackingnet,
  title={Trackingnet: A large-scale dataset and benchmark for object tracking in the wild},
  author={Muller, Matthias and Bibi, Adel and Giancola, Silvio and Alsubaihi, Salman and Ghanem, Bernard},
  booktitle={ECCV},
  pages={300--317},
  year={2018}
}

@inproceedings{fan2019lasot,
  title={Lasot: A high-quality benchmark for large-scale single object tracking},
  author={Fan, Heng and Lin, Liting and Yang, Fan and Chu, Peng and Deng, Ge and Yu, Sijia and Bai, Hexin and Xu, Yong and Liao, Chunyuan and Ling, Haibin},
  booktitle={CVPR},
  pages={5374--5383},
  year={2019}
}

@article{huang2019got,
  title={Got-10k: A large high-diversity benchmark for generic object tracking in the wild},
  author={Huang, Lianghua and Zhao, Xin and Huang, Kaiqi},
  journal={TPAMI},
  year={2019},
  publisher={IEEE}
}

@InProceedings{Wu_2023_CVPR_dropmae,
    author    = {Wu, Qiangqiang and Yang, Tianyu and Liu, Ziquan and Wu, Baoyuan and Shan, Ying and Chan, Antoni B.},
    title     = {DropMAE: Masked Autoencoders With Spatial-Attention Dropout for Tracking Tasks},
    booktitle = {Proceedings of the IEEE/CVF Conference on Computer Vision and Pattern Recognition (CVPR)},
    month     = {June},
    year      = {2023},
    pages     = {14561-14571}
}

@inproceedings{lin2014microsoft,
  title={Microsoft coco: Common objects in context},
  author={Lin, Tsung-Yi and Maire, Michael and Belongie, Serge and Hays, James and Perona, Pietro and Ramanan, Deva and Doll{\'a}r, Piotr and Zitnick, C Lawrence},
  booktitle={ECCV},
  pages={740--755},
  year={2014},
}

@ARTICLE{otb2015,
  author={Y. {Wu} and J. {Lim} and M. {Yang}},
  journal={TPAMI},
  title={Object Tracking Benchmark},
  year={2015},
  volume={37},
  number={9},
  pages={1834--1848},
  doi={10.1109/TPAMI.2014.2388226}}

@inproceedings{mueller2016benchmark,
  title={A benchmark and simulator for uav tracking},
  author={Mueller, Matthias and Smith, Neil and Ghanem, Bernard},
  booktitle={ECCV},
  pages={445--461},
  year={2016},
}
}

\end{document}